\documentclass[10pt,twocolumn,letterpaper]{article}

\usepackage{cvpr}
\usepackage{times}
\usepackage{epsfig}
\usepackage{graphicx}
\usepackage{amsmath}
\usepackage{amssymb}

\usepackage{capt-of}
\usepackage{dblfloatfix}
\usepackage{booktabs}
\usepackage{multirow}
\usepackage{tabularx}
\newcolumntype{Y}{>{\centering\arraybackslash}X}
\usepackage{bm}
\newtheorem{assumption}{Assumption}
\usepackage{breakcites}

\usepackage[pagebackref=true,breaklinks=true,letterpaper=true,colorlinks,bookmarks=false]{hyperref}

\cvprfinalcopy

\def\cvprPaperID{****} 

\begin{document}

\title{\vspace{-.7em}Noise Robust Generative Adversarial Networks\vspace{-.7em}}

\author{
  Takuhiro Kaneko$^1$
  \quad Tatsuya Harada$^{1,2}$
  \vspace{3mm}\\
  $^1$The University of Tokyo
  \quad $^2$RIKEN
  \vspace{-1mm}
}

\twocolumn[{
  \renewcommand\twocolumn[1][]{#1}
  \maketitle
  \vspace{-6mm}
  \begin{center}
    \includegraphics[width=\textwidth]{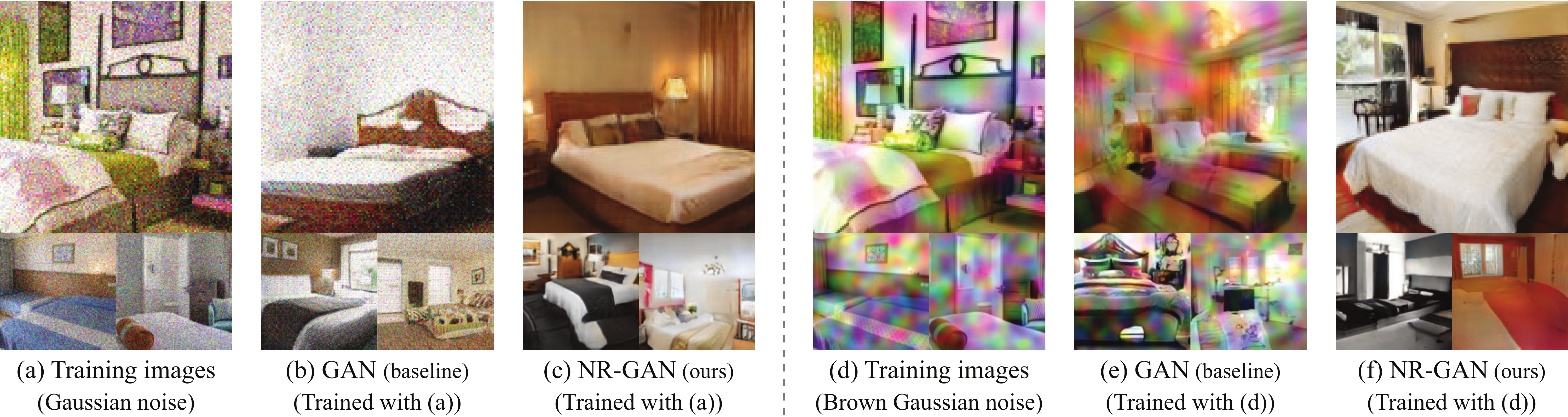}
  \end{center}
  \vspace{-3mm}
  \captionof{figure}{
    \textbf{Examples of noise robust image generation.} Recent GANs have shown promising results for reproducing training images. However, even when training images are noisy (a)(d), they attempt to reproduce the training images faithfully, as shown in (b)(e). To remedy this, we propose noise robust GANs (NR-GANs), which can learn to generate clean images (c)(f), even when training images are noisy (a)(d). Our NR-GANs are unique in that they solve this problem without full knowledge of the noise (e.g., the noise distribution type, noise amount, or signal-noise relationship). Indeed, in (c) and (f), although the same models (in particular, SI-NR-GAN-II, which is a variant of NR-GANs) are used for different noises (a)(d), they succeed in learning clean image generators adaptively through training.
  }
  \label{fig:concept}
  \vspace{4mm}
}]

\begin{abstract}
  \vspace{-3mm}
  Generative adversarial networks (GANs) are neural networks that learn data distributions through adversarial training. In intensive studies, recent GANs have shown promising results for reproducing training images. However, in spite of noise, they reproduce images with fidelity. As an alternative, we propose a novel family of GANs called noise robust GANs (NR-GANs), which can learn a clean image generator even when training images are noisy. In particular, NR-GANs can solve this problem without having complete noise information (e.g., the noise distribution type, noise amount, or signal-noise relationship). To achieve this, we introduce a noise generator and train it along with a clean image generator. However, without any constraints, there is no incentive to generate an image and noise separately. Therefore, we propose distribution and transformation constraints that encourage the noise generator to capture only the noise-specific components. In particular, considering such constraints under different assumptions, we devise two variants of NR-GANs for signal-independent noise and three variants of NR-GANs for signal-dependent noise. On three benchmark datasets, we demonstrate the effectiveness of NR-GANs in noise robust image generation. Furthermore, we show the applicability of NR-GANs in image denoising. Our code is available at \url{https://github.com/takuhirok/NR-GAN/}.
  \vspace{-4mm}
\end{abstract}

\begin{figure*}[t]
  \centering
  \includegraphics[width=1.0\textwidth]{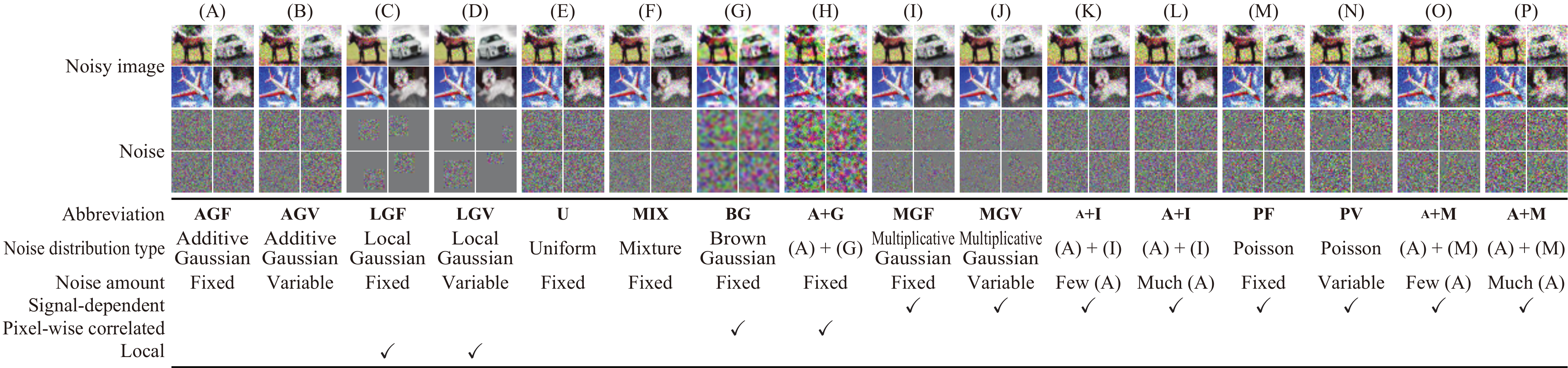}
  \vspace{-4mm}
  \caption{
    \textbf{Noise categorization and examples.} In this paper, we handle various types of noise. (A)(B) Additive Gaussian noise with fixed ${\sigma} = 25$ (A) and variable ${\sigma} \in [5, 50]$ (B), where ${\sigma}$ is the standard deviation of a Gaussian distribution. (C)(D) Local Gaussian noise with fixed patch $p_h = p_w = 16$ (C) and variable patch $p_h, p_w \in [8, 24]$ (D), where $p_h$ and $p_w$ are the height and width of noise region, respectively. ${\sigma}$ is fixed to 25 in both cases. (E) Uniform noise $[-50, 50]$. (F) Mixture noise~\cite{QZhaoICML2014}. Mixture of 10\% uniform noise $[-50, 50]$, 20\% Gaussian noise ($\sigma = 25$), and 70\% Gaussian noise ($\sigma = 15$). (G) Brown Gaussian noise~\cite{JLehtinenICML2018}. A Gaussian filter (kernel size is $5 \times 5$) is applied to (A). (H) Additive Brown Gaussian noise. (G) is added to (A) while remaining the original (A). (I)(J) Multiplicative Gaussian noise with fixed ${\sigma} = 25$ (I) and variable ${\sigma} \in [5, 50]$ (J). (K)(L) Additive and multiplicative Gaussian noise. (K) Sum of few (A) (${\sigma} = 5$) and (I). (L) Sum of much (A) (${\sigma} = 25$) and (I). (M)(N) Poisson noise with fixed ${\lambda} = 30$ (M) and variable ${\lambda} \in [10, 50]$ (N), where $\lambda$ is the total number of events. (O)(P) Poisson-Gaussian noise. (O) Sum of few (A) (${\sigma} = 5$) and (M). (P) Sum of much (A) (${\sigma} = 25$) and (M).
  }
  \label{fig:noise_examples}
  \vspace{-5mm}
\end{figure*}

\section{Introduction}
\label{sec:introduction}

In computer vision and machine learning, generative models have been actively studied and used to generate or reproduce an image that is indistinguishable from a real image. Generative adversarial networks (GANs)~\cite{IGoodfellowNIPS2014}, which learn data distributions through adversarial training, have garnered special attention owing to their ability to produce high-quality images. In particular, with recent advancements~\cite{MArjovskyICML2017,XMaoICCV2017,JHLimArXiv2017,IGulrajaniNIPS2017,NKodaliArXiv2017,LMeschederICML2018,TMiyatoICLR2018b,TKarrasICLR2018,HZhangICML2019}, the latest GANs (e.g., BigGAN~\cite{ABrockICLR2019} and StyleGAN~\cite{TKarrasCVPR2019}) have succeeded in generating images indistinguishable for humans.

However, a persistent issue is that recent high-capacity GANs could replicate images faithfully even though the training images were noisy. Indeed, as shown in Figure~\ref{fig:concept}(b)(e), when standard GAN is trained with noisy images, it attempts to recreate them. Although the long-term development of devices has steadily improved image quality, image degradation is unavoidable in real situations. For example, electronic noise is inevitable in digital imaging~\cite{TPlotzCVPR2017,JAnayaVCIR2018} and estimator variance often appears as noise in graphic rendering~\cite{MZwickerCGF2015,BXuTOG2019}. Therefore, susceptibility to noise is practically undesirable for GANs.

The question becomes: \textit{``How can we learn a clean image generator even when only noisy images are available for training?''} We call this problem \textit{noise robust image generation}. One solution is to apply a denoiser as pre-process. However, a limitation is that the generator performance highly relies on the quality of the denoiser, which is relatively difficult to learn when clean images are not available for training. As an alternative, AmbientGAN~\cite{ABoraICLR2018} was recently proposed, which provides a promising solution by simulating the noise corruption on the generated images and learning the discriminator so that it distinguishes a real noisy image from a simulatively corrupted generated image. This makes it possible to learn a clean image generator directly from noisy images without relying on a denoiser.

However, a key limitation of AmbientGAN is that it assumes that the noise corruption process is pre-defined. Therefore, to utilize it, we need to have all information about the noise, such as the noise distribution type (e.g., Gaussian), noise amount (e.g., standard deviation), and signal-noise relationship. For instance, to treat 16 noises shown in Figure~\ref{fig:noise_examples}, we need to carefully prepare 16 noise simulation models that depend on the noise.

To deal with this, we propose \textit{noise robust GANs} (\textit{NR-GANs}), which can achieve noise robust image generation without having complete noise information. Our main idea is as follows. We first introduce two generators, a clean image generator and noise generator. To make them generate an image and noise, respectively, we impose a distribution or transformation constraint on the noise generator so that it only captures the components that follow the specified distribution or transformation invariance. As such a constraint can take various forms depending on the type of assumptions; we develop five variants: two \textit{signal-independent NR-GANs} (\textit{SI-NR-GANs}) and {three \textit{signal-dependent NR-GANs} (\textit{SD-NR-GANs}). Figure~\ref{fig:concept}(c)(f) shows examples of images generated using NR-GANs. Here, although the same models are used for different noises (a)(d), NR-GANs succeed in learning clean image generators adaptively.

As the noise robustness of GANs has not been sufficiently studied, we first perform a comprehensive study on \textsc{CIFAR-10}~\cite{AKrizhevskyTech2009}, where we compare various models in diverse noise settings (in which we test 152 conditions). Furthermore, inspired by the recent large-scale study on GANs~\cite{KKurachICML2019}, we also examin the performance on more complex datasets (\textsc{LSUN Bedroom}~\cite{FYuArXiv2015} and \textsc{FFHQ}~\cite{TKarrasCVPR2019}). Finally, we demonstrate the applicability of NR-GANs in image denoising, where we learn a denoiser using generated noisy images and generated clean images (GN2GC), and empirically examine a chicken and egg problem between noise robust image generation and image denoising.

Overall, our contributions are summarized as follows:
\begin{itemize}
  \vspace{-1.5mm}
  \setlength{\parskip}{0.75pt}
  \setlength{\itemsep}{0.75pt}
\item
  We provide \textit{noise robust image generation}, the purpose of which is to learn a clean image generator even when training images are noisy. In particular, we solve this problem without full knowledge of the noise.
\item
  To achieve this, we propose a novel family of GANs called \textit{NR-GANs} that train a clean image generator and noise generator simultaneously with a distribution or transformation constrain on the noise generator.
\item
  We provide a comprehensive study on \textsc{CIFAR-10} (in which we test 152 conditions) and examine the versatility in more complex datasets (\textsc{LSUN Bedroom} and \textsc{FFHQ}); finally, we demonstrate the applicability in image denoising. The project page is available at \url{https://takuhirok.github.io/NR-GAN/}.
\end{itemize}

\section{Related work}
\label{sec:related_work}

\smallskip\noindent\textbf{Deep generative models.}
Image generation is a fundamental problem and has been intensively studied in computer vision and machine learning. Recently, deep generative models have emerged as a promising framework. Among them, three prominent models along with GANs are variational autoencoders~\cite{DKingmaICLR2014,DRezendeICML2014}, autoregressive models~\cite{AOordICML2016}, and flow-based models~\cite{LDinhICLRW2015,LDinhICLR2017}. Each model has pros and cons. A well-known disadvantage of GANs is training instability; however, it has been steadily improved by recent advancements~\cite{MArjovskyICML2017,XMaoICCV2017,JHLimArXiv2017,MBellemarearXiv2017,TSalimansICLR2018,IGulrajaniNIPS2017,NKodaliArXiv2017,LMeschederICML2018,TMiyatoICLR2018b,TKarrasICLR2018,HZhangICML2019,ABrockICLR2019,TChenICLR2019,TKarrasCVPR2019}. In this work, we focus on GANs for their design flexibility, which allows them to incorporate the core of our models, a noise generator and its constraints. Also in other models, image fidelity has improved~\cite{AOordNIPS2017,ARazaviNeurIPS2019,JMenickICLR2019,DKingmaNeurIPS2018}. Hence, sensitivity to noise can be problematic. Incorporating our ideas into them is a possible direction of future work.

\smallskip\noindent\textbf{Image denoising.}
Image denoising is also a fundamental problem and several methods have been proposed. They are roughly categorized into two: model-based methods~\cite{KDabovTIP2007,SGuCVPR2014,MEladTIP2006,JMairalTIP2007,ABuadesCVPR2005,JMairalICCV2009,WDongTIP2012,FLuisierTIP2010,MMakitaloTIP2012} and discriminative learning methods~\cite{VJainNIPS2009,XMaoNIPS2016,YTaiICCV2017,KZhangTIP2017,KZhangTIP2018,JChenCVPR2018,SGuoCVPR2019,JLehtinenICML2018,AKrullCVPR2019,JBatsonICML2019}. Recently, discriminative learning methods have shown a better performance; however, a limitation is that most of them (i.e., Noise2Clean (N2C)) require clean images for supervised training of a denoiser. To handle this, self-supervised learning methods (e.g., Noise2Void (N2V) \cite{AKrullCVPR2019} and Noise2Self (N2S) \cite{JBatsonICML2019}) were proposed. These methods assume the same data setting as ours, i.e., only noisy images are available for training. However, they still have some limitations, e.g., they cannot handle pixel-wise correlated noise, such as shown in Figure~\ref{fig:noise_examples}(G)(H), and their performance is still inferior to supervised learning methods.

Image denoising and our noise robust image generation is a chicken and egg problem and each task can be used as a pre-task for learning the other. In the spirit of AmbientGAN, we aim to learn a clean image generator directly from noisy images. However, examining the performance on (1) learning a generator using denoised images and (2) learning a denoiser using generated clean and noisy images is an interesting research topic. Motivated by this, we empirically examined them through comparative studies. We provide the results in Sections~\ref{subsec:comprehensive_study} and \ref{subsec:application}.

\smallskip\noindent\textbf{Noise robust models.}
Except for image denoising, noise robust models have been studied in image classification to learn a classifier in practical settings. There are two studies addressing label noise~\cite{AGhoshAAAI2017,ZZhangNeurIPS2018,SReedICLR2015,EMalachNIPS2017,LJiangICML2018,DTanakaCVPR2018,MRenICML2018,BHanNeurIPS2018,SSukhbaatarICLRW2015,IJindalICDM2016,GPatriniCVPR2017,JGoldbergerICLR2017} and addressing image noise~\cite{SZhengCVPR2016,SDiamondArXiv2017}. For both tasks, the issue is the memorization effect~\cite{CZhangICLR2017}, i.e., DNN classifiers can fit labels or images even though they are noisy or fully corrupted. As demonstrated in Figure~\ref{fig:concept}, a similar issue also occurs in image generation.

Pertaining image generation, handling of label noise~\cite{TKanekoCVPR2019,TKanekoBMVC2019,TKanekoArXiv2019a,KKThekumparampilNeurIPS2018} and image noise~\cite{ABoraICLR2018} has begun to be studied. Our NR-GANs are categorized into the latter. As discussed in Section~\ref{sec:introduction}, AmbientGAN~\cite{ABoraICLR2018} is a representative model in the latter category. However, a limitation is that it requires full knowledge of the noise. Therefore, we introduce NR-GANs to solve this problem as they do not have this limitation.

\section{Notation and problem statement}
\label{sec:notation}

We first define notation and the problem statement. Hereafter, we use superscripts $r$ and $g$ to denote the real distribution and generative distribution, respectively. Let ${\bm y}$ be the observable noisy image and ${\bm x}$ and ${\bm n}$ be the underlying signal (i.e., clean image) and noise, respectively, where ${\bm y}, {\bm x}, {\bm n} \in \mathbb{R}^{H \times W \times C}$ ($H$, $W$, and $C$ are the height, width, and channels of an image, respectively). In particular, we assume that ${\bm y}$ can be decomposed additively: ${\bm y} = {\bm x} + {\bm n}$.\footnote{We decompose additively; however, note that this representation includes signal-independent noise ${\bm n} \sim p({\bm n})$ and signal-dependent noise ${\bm n} \sim p({\bm n}|{\bm x})$.} Our task is to learn a clean image generator that can reproduce clean images, such that $p^g({\bm x}) = p^r({\bm x})$, when trained with noisy images ${\bm y}^r \sim p^r({\bm y})$. This is a challenge for standard GAN as it attempts to mimic the observable images including the noise; namely, it learns $p^g({\bm y}) = p^r({\bm y})$.

We assume various types of noise. Figure~\ref{fig:noise_examples} shows the categorization and examples of the noises that we address in this paper. They include signal-independent noises (A)--(H), signal-dependent noises (I)--(P), pixel-wise correlated noises (G)(H), local noises (C)(D), and their combination (H)(K)(L)(O)(P). We also consider two cases: the noise amount is either is fixed or variable across the dataset.

As discussed in Section~\ref{sec:introduction}, one solution is AmbientGAN~\cite{ABoraICLR2018}; however, it is limited by the need for prior noise knowledge. We plan a solution that will not require that full prior knowledge. Our central idea is to introduce two generators, i.e., a clean image generator and noise generator, and impose a distribution or transformation constraint on the noise generator so that it captures only the noise-specific components. In particular, we explicate such constraints by relying on the signal-noise dependency. We first review our baseline AmbientGAN~\cite{ABoraICLR2018} (Section~\ref{sec:AmbientGAN}); then detail NR-GANs for signal-independent noise (Section~\ref{sec:SI-NR-GAN}) and signal-dependent noise (Section~\ref{sec:SD-NR-GAN}).

\begin{figure*}[t]
  \centering
  \includegraphics[width=1.0\textwidth]{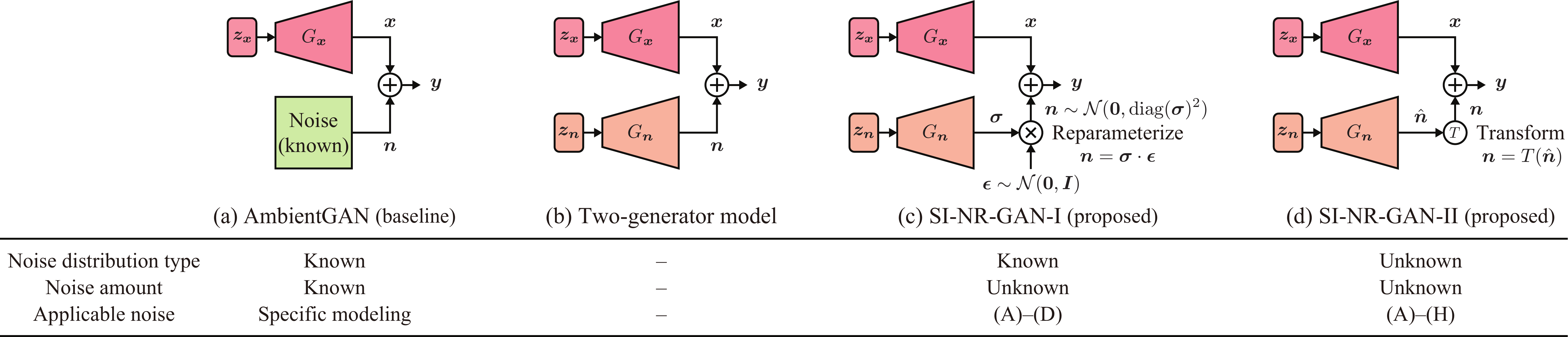}
  \vspace{-4mm}
  \caption{
    \textbf{Comparison of AmbientGAN (baseline) and SI-NR-GANs (proposed).} Because the discriminators are the same, we only depict the generators. (a) AmbientGAN assumes that the noise model is pre-defined. (b) To mitigate this requirement, we introduce a two-generator model and learn a noise generator $G_{\bm n}$ along with a clean image generator $G_{\bm x}$. (c) To make $G_{\bm n}$ capture only the noise specific components, SI-NR-GAN-I regularizes the output distribution of $G_{\bm n}$ using a reparameterization trick under the assumption that the noise distribution type is known. (d) Furthermore, considering the situation when the noise distribution type is unknown, we develop SI-NR-GAN-II, which applies transformations ${\bm n} = T(\hat{\bm n})$ to extract the transformation-invariant element, i.e., noise.
  }
  \label{fig:models_si}
  \vspace{-4mm}
\end{figure*}

\section{Baseline: AmbientGAN}
\label{sec:AmbientGAN}

AmbientGAN~\cite{ABoraICLR2018} (Figures~\ref{fig:models_si}(a) and \ref{fig:models_sd}(a)) is a variant of GANs, which learns an underlying distribution $p^r({\bm x})$ only from noisy images ${\bm y}^r \sim p^r({\bm y})$.\footnote{Strictly, AmbientGAN can handle more general \textit{lossy} data, such as missing data. Here, we narrow the target in accordance with our task.} This is a challenging because the desired images ${\bm x}^r \sim p^r({\bm x})$ are not observable during training. To overcome this challenge, AmbientGAN introduces a noise simulation model ${\bm y} = F_{\bm \theta}({\bm x})$ under the assumption that it is priorly known. The main idea of AmbientGAN is to incorporate this noise simulation model into the adversarial training framework:
\begin{flalign}
  \label{eqn:AmbientGAN}
  & \min_{G_{\bm x}} \max_{D_{\bm y}} \mathbb{E}_{{\bm y}^r \sim p^r({\bm y})} [\log D_{\bm y}({\bm y}^r)]
  \nonumber \\
  & \: + \mathbb{E}_{{\bm z}_{\bm x} \sim p({\bm z}_{\bm x}), {\bm \theta} \sim p({\bm \theta})} [\log(1 - D_{\bm y}(F_{\bm \theta}(G_{\bm x}({\bm z}_{\bm x}))))].
\end{flalign}
Just like standard GAN, a generator $G_{\bm x}$ transforms the latent vector ${\bm z}_{\bm x}$ into an image ${\bm x}^g = G_{\bm x}({\bm z}_{\bm x})$. However, differently from the standard GAN discriminator, which directly distinguishes a real image ${\bm y}^r$ from a generated image ${\bm x}^g$, the AmbientGAN discriminator $D_{\bm y}$ distinguishes ${\bm y}^r$ from a \textit{noised} generated image ${\bm y}^g = F_{\bm \theta}({\bm x}^g)$. Intuitively, this modification allows noisy $p^g({\bm y})$ ($= p^g(F_{\bm \theta}({\bm x}))$) to get close to noisy $p^r({\bm y})$ ($= p^r(F_{\bm \theta}({\bm x}))$). When $F_{\bm \theta}$ is invertible or uniquely determined, underlying clean $p^g({\bm x})$ also approaches underlying clean $p^r({\bm x})$.

\section{Signal-independent noise robust GANs}
\label{sec:SI-NR-GAN}

As described above, a limitation of AmbientGAN is that it requires that a noise simulation model $F_{\bm \theta}({\bm x})$ is priorly known. To alleviate this, we introduce a noise generator ${\bm n} = G_{\bm n}({\bm z}_{\bm n})$ (Figure~\ref{fig:models_si}(b)) and train it along with a clean image generator $G_{\bm x}$ using the following objective function:
\begin{flalign}
  \label{eqn:NR-GAN}
  & \min_{G_{\bm x}, G_{\bm n}} \max_{D_{\bm y}} \mathbb{E}_{{\bm y}^r \sim p^r({\bm y})} [\log D_{\bm y}({\bm y}^r)]
  \nonumber \\
  & \: + \mathbb{E}_{{\bm z}_{\bm x} \sim p({\bm z}_{\bm x}), {\bm z}_{\bm n} \sim p({\bm z}_{\bm n})} [\log(1 - D_{\bm y}(G_{\bm x}({\bm z}_{\bm x}) + G_{\bm n}({\bm z}_{\bm n})))].
\end{flalign}
Nevertheless, without any constraints, there is no incentive to make $G_{\bm x}$ and $G_{\bm n}$ generate an image and a noise, respectively. Therefore, we provide a constraint on $G_{\bm n}$ so that it captures only the noise-specific components. In particular, we develop two variants that have different assumptions: SI-NR-GAN-I (Section~\ref{subsec:SI-NR-GAN-I}) and SI-NR-GAN-II (Section~\ref{subsec:SI-NR-GAN-II}).

\begin{figure*}[t]
  \centering
  \includegraphics[width=1.0\textwidth]{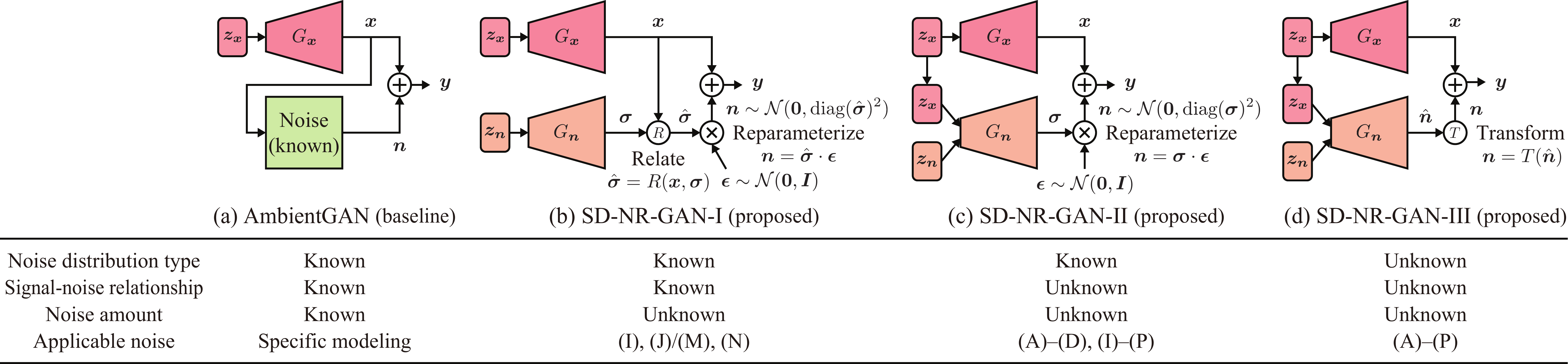}
  \vspace{-4mm}
  \caption{
    \textbf{Comparison of AmbientGAN (baseline) and SD-NR-GANs (proposed).} Because the discriminators are the same, we only depict the generators. (a) AmbientGAN pre-defines the noise model. (b) SD-NR-GAN-I represents the signal-noise relationship explicitly, while the noise amount is estimated through training. (c) SD-NR-GAN-II expresses the signal-noise relationship implicitly and this relationship and the noise amount are acquired through training. (d) SD-NR-GAN-III only imposes a weak constraint via the transformation and learns the noise distribution type, signal-noise relationship, and noise amount through training.
  }
  \label{fig:models_sd}
  \vspace{-4.5mm}
\end{figure*}

\subsection{SI-NR-GAN-I}
\label{subsec:SI-NR-GAN-I}

In SI-NR-GAN-I, we assume the following.
\begin{assumption}
  \label{assumption:SI-NR-GAN-I}
  (i) The noise ${\bm n}$ is conditionally pixel-wise independent given the signal ${\bm x}$.
  (ii) The noise distribution type (e.g., Gaussian) is priorly known. Note that the noise amount needs not to be known.
  (iii) The signal ${\bm x}$ does not follow the defined noise distribution.  
\end{assumption}
Under this assumption, we develop SI-NR-GAN-I (Figure~\ref{fig:models_si}(c)). In this model, we regularize the output distribution of $G_{\bm n}$ in a pixel-wise manner using a reparameterization trick~\cite{DKingmaICLR2014}. Here, we present the case when the noise distribution type is defined as zero-mean Gaussian:\footnote{Strictly, our approach is applicable as long as a noise follows a differentiable distribution \cite{DKingmaICLR2014}.}
\begin{flalign}
  \label{eqn:additive_gaussian}
  {\bm y} = {\bm x} + {\bm n}, \text{where } {\bm n} \sim {\cal N}({\bm 0}, \text{diag}({\bm \sigma})^2),
\end{flalign}
where ${\bm \sigma}  \in {\mathbb{R}^{H \times W \times C}}$ is the pixel-wise standard deviation. In this case, we redefine the noise generator as ${\bm \sigma} = G_{\bm n}({\bm z}_{\bm n})$; and introduce an auxiliary pixel-wise random variable ${\bm \epsilon} \sim {\cal N}({\bm 0}, {\bm I})$, where ${\bm \epsilon} \in {\mathbb{R}^{H \times W \times C}}$; and then calculate the noise ${\bm n}$ by multiplying them: ${\bm n} = {\bm \sigma} \cdot {\bm \epsilon}$, where $\cdot$ represents an element-wise product. This formulation allows the noise to be sampled as ${\bm n} \sim {\cal N}({\bm 0}, \text{diag}({\bm \sigma})^2)$.

In SI-NR-GAN-I, ${\bm \sigma}$ is learned through training in a pixel-wise manner. Therefore, the same model can be applied to various noises (e.g., Figure~\ref{fig:noise_examples}(A)--(D), in which each pixel's noise follows a Gaussian distribution, while the noise amount is different in a sample-wise (e.g., (B)) or pixel-wise (e.g., (D)) manner).

\subsection{SI-NR-GAN-II}
\label{subsec:SI-NR-GAN-II}

Two limitations of SI-NR-GAN-I are that it assumes that (i) the noise is pixel-wise independent and (ii) the noise distribution type is pre-defined. The first assumption makes it difficult to apply to a pixel-wise correlated noise (e.g., Figure~\ref{fig:noise_examples}(G)(H)). The second assumption could cause difficulty when diverse noises are mixed (e.g., Figure~\ref{fig:noise_examples}(F)) or the noise distribution type is different from the pre-defined (e.g., Figure~\ref{fig:noise_examples}(E)). This motivates us to devise SI-NR-GAN-II, which works under a different assumption:
\begin{assumption}
  \label{assumption:SI-NR-GAN-II}
  (i) The noise ${\bm n}$ is rotation-, channel-\hbox{shuffle-,} or color-inverse-invariant.
  (ii) The signal ${\bm x}$ is rotation-, channel-shuffle-, or color-inverse-variant.
\end{assumption}
Among the noises in Figure~\ref{fig:noise_examples}, this assumption holds in all signal-independent noises (A)--(H). This assumption is reasonable when ${\bm n}$ is a zero-mean signal-independent noise and ${\bm x}$ is a natural image.\footnote{In fact, these kinds of transformations (especially, rotation) are commonly used in self-supervised learning~\cite{SGidarisICLR2018,AKolesnikovCVPR2019,TChenCVPR2019,MLucicICML2019}, which utilize the transformations to learn natural image representations. Inspired by this, we employ the transformations to isolate noises from natural images.} Under this assumption, we establish SI-NR-GAN-II (Figure~\ref{fig:models_si}(d)). In this model, we redefine the noise generator as $\hat{\bm n} = G_{\bm n}({\bm z}_{\bm n})$ ($\hat{\bm n} \in \mathbb{R}^{H \times W \times C}$) and apply transformations to $\hat{\bm n}$ by ${\bm n} = T(\hat{\bm n})$, where $T$ is a transformation function. As $T$, we can use arbitrary transformation as long as it is applicable to ${\bm n}$ but not allowable to ${\bm x}$. In practice, we use three transformations: (i) \textit{rotation} -- rotating $\hat{\bm n}$ by $d \in \{ 0^{\circ}, 90^{\circ}, 180^{\circ}, 270^{\circ} \}$ randomly, (ii) \textit{channel shuffle} -- shuffling RGB channels randomly, and (iii) \textit{color inversion} -- inverting colors randomly in a channel-wise manner. Each one utilizes one of the invariant and variant characteristics mentioned in Assumption~\ref{assumption:SI-NR-GAN-II}. In SI-NR-GAN-II, the noise origin $\hat{\bm n}$ is acquired in a data-driven manner; therefore, it is applicable to diverse noises (e.g., Figure~\ref{fig:noise_examples}(A)--(H)) without model modifications.

\section{Signal-dependent noise robust GANs}
\label{sec:SD-NR-GAN}

Just like in the signal-independent noise case, AmbientGAN is applicable to signal-dependent noise by incorporating the pre-defined noise model (Figure~\ref{fig:models_sd}(a)). However, it requires prior knowledge about the noise distribution type, signal-noise relationship, and noise amount. To deal with these requirements, we establish three variants that have different assumptions: SD-NR-GAN-I (Section~\ref{subsec:SD-NR-GAN-I}), SD-NR-GAN-II (Section~\ref{subsec:SD-NR-GAN-II}), and SD-NR-GAN-III (Section~\ref{subsec:SD-NR-GAN-III}).

\subsection{SD-NR-GAN-I}
\label{subsec:SD-NR-GAN-I}

We first consider the case when the following assumption holds in addition to Assumption~\ref{assumption:SI-NR-GAN-I}.
\begin{assumption}
  \label{assumption:SD-NR-GAN-I}
  The signal-noise relationship is priorly known. Note that the noise amount needs not be known.
\end{assumption}
Under this assumption, we devise SD-NR-GAN-I (Figure~\ref{fig:models_sd}(b)), which incorporates a signal-noise relational procedure into SI-NR-GAN-I \textit{explicitly}. In particular, we devise two configurations for two typical signal-dependent noises: multiplicative Gaussian noise (Figure~\ref{fig:noise_examples}(I)(J)) and Poisson noise (Figure~\ref{fig:noise_examples}(M)(N)).

Multiplicative Gaussian noise is defined as
\begin{flalign}
  \label{eqn:multiplicative_gaussian}
  {\bm y} = {\bm x} + {\bm n}, \text{where } {\bm n} \sim {\cal N}({\bm 0}, \text{diag}({\bm \sigma} \cdot {\bm x})^2).
\end{flalign}
To represent this noise with trainable ${\bm \sigma}$, we redesign the noise generator as ${\bm \sigma} = G_{\bm n}({\bm z}_{\bm n})$. Then, we convert ${\bm \sigma}$ using a signal-noise relational function $R({\bm x}, {\bm \sigma}) = {\bm \sigma} \cdot {\bm x} = \hat{\bm \sigma}$. Finally, we obtain ${\bm n} \sim {\cal N}({\bm 0}, \text{diag}(\hat{\bm \sigma})^2)$ by using the reparameterization trick described in Section~\ref{subsec:SI-NR-GAN-I}.

Poisson noise (or shot noise) is sampled by ${\bm y} \sim \text{Poisson}({\bm \lambda} \cdot {\bm x}) / {\bm \lambda}$, where ${\bm \lambda}$ is the total number of events. As this noise is discrete and intractable to construction of a differentiable model, we use a Gaussian approximation~\cite{SWHasinoff2014}, which is commonly used for Poisson noise modeling:
\begin{flalign}
  \label{eqn:approx_poisson}
  {\bm y} = {\bm x} + {\bm n}, \text{where } {\bm n} \sim {\cal N}({\bm 0}, \text{diag}({\bm \sigma} \cdot \sqrt{\bm x})^2),
\end{flalign}
where ${\bm \sigma} = \sqrt{1 / {\bm \lambda}}$. The implementation method is almost the same as that for the multiplicative Gaussian noise except that we redefine $R({\bm x}, {\bm \sigma})$ as $R({\bm x}, {\bm \sigma}) = {\bm \sigma} \cdot \sqrt{{\bm x}} = \hat{\bm \sigma}$.

In both noise cases, the noise amount ${\bm \sigma}$ is trainable; therefore, each configuration of SD-NR-GAN-I is applicable to the noises in Figure~\ref{fig:noise_examples}(I)(J) and those in Figure~\ref{fig:noise_examples}(M)(N), respectively, without model modifications.

\subsection{SD-NR-GAN-II}
\label{subsec:SD-NR-GAN-II}

In SD-NR-GAN-II, we consider the case when the noise distribution type is known (i.e., Assumption~\ref{assumption:SI-NR-GAN-I} holds) but the signal-noise relationship is unknown (i.e., Assumption~\ref{assumption:SD-NR-GAN-I} is not required). Under this assumption, we aim to learn $R({\bm x}, {\bm \sigma})$ \textit{implicitly}, which is explicitly given in SD-NR-GAN-I. To achieve this, we develop SD-NR-GAN-II (Figure~\ref{fig:models_sd}(c)), which is an extension of SI-NR-GAN-I incorporating the image latent vector ${\bm z}_{\bm x}$ into the input of $G_{\bm n}$, i.e., ${\bm \sigma} = G_{\bm n}({\bm z}_{\bm n}, {\bm z}_{\bm x})$. Similarly to SI-NR-GAN-I, we sample ${\bm n} \sim {\cal N}({\bm 0}, \text{diag}({\bm \sigma})^2)$ using the reparameterization trick described in Section~\ref{subsec:SI-NR-GAN-I}. Here, we consider the case when the noise distribution type is defined as zero-mean Gaussian.

As discussed in Section~\ref{subsec:SD-NR-GAN-I}, multiplicative Gaussian noise and Poisson noise are represented (or approximated) as signal-dependent Gaussian noise; therefore, SD-NR-GAN-II is applicable to these noises (e.g., Figure~\ref{fig:noise_examples}(I)(J)(M)(N)). Furthermore, SD-NR-GAN-II can internally learn $R({\bm x}, {\bm \sigma})$; therefore, the same model can also be applied to signal-independent noise (Figure~\ref{fig:noise_examples}(A)--(D)), i.e., $R({\bm x}, {\bm \sigma}) = {\bm x}$, and the combination of multiple noises (Figure~\ref{fig:noise_examples}(K)(L)(O)(P)), e.g., $R({\bm x}, {\bm \sigma}_d, {\bm \sigma}_i) = {\bm \sigma}_d \cdot {\bm x} + {\bm \sigma}_i$.

\subsection{SD-NR-GAN-III}
\label{subsec:SD-NR-GAN-III}

Finally, we deal with the case when both the noise distribution type and signal-noise relationship are not known. In this case, we impose a similar assumption as Assumption~\ref{assumption:SI-NR-GAN-II}. However, \textit{rotation} and \textit{channel shuffle} collapse the per-pixel signal-noise dependency that is included in typical signal-dependent noise (e.g., Figure~\ref{fig:noise_examples}(I)--(P)). Therefore, we only induce the assumption regarding \textit{color inversion}.\footnote{For further clarification, we conducted a comparative study on transformations in SD-NR-GAN-III. See Appendix~\ref{subsec:comparative_transformations} for details.} Under this assumption, we devise SD-NR-GAN-III (Figure~\ref{fig:models_sd}(d)). Similarly to SD-NR-GAN-II, SD-NR-GAN-III learns the signal-noise relationship \textit{implicitly} by incorporating ${\bm z}_{\bm x}$ into the input of $G_{\bm n}$, i.e., $\hat{\bm n} = G_{\bm n}({\bm z}_{\bm n}, {\bm z}_{\bm x})$. Similarly to SI-NR-GAN-II, we impose a transformation constraint on $G_{\bm n}$ by applying ${\bm n} = T(\hat{\bm n})$, where $T$ is defined as \textit{color inversion}. The noise origin $\hat{\bm n}$ is learned through training; therefore, SD-NR-GAN-III can be adopted to various noises (e.g., all noises in Figure~\ref{fig:noise_examples}) without modifying the model.

\section{Advanced techniques for practice}
\label{sec:advanced_techniques}

\subsection{Alleviation of convergence speed difference}
\label{subsec:convergence_speed_difference}

In proposed NR-GANs, $G_{\bm x}$ and $G_{\bm n}$ are learned simultaneously. Ideally, we expect that $G_{\bm x}$ and $G_{\bm n}$ would be optimized at the same speed; however, through experiments, we found that $G_{\bm n}$ tends to learn faster than $G_{\bm x}$ and results in a mode collapse in the early training phase. A possible cause is that the noise distribution is simpler and easier to learn than the image distribution. To address this problem, we apply the diversity-sensitive regularization~\cite{DYangICLR2019} to $G_{\bm n}$. Intuitively, this regularization makes $G_{\bm n}$ sensitive to ${\bm z}_{\bm n}$ and has an effect to prevent the mode collapse. In the experiments, we incorporate this technique to all NR-GANs. We discuss the effect of this regularization in Appendix~\ref{subsec:effect_ds_regularization}.

\subsection{Alleviation of approximation degradation}
\label{subsec:approximation_degradation}

As described in Section~\ref{subsec:SD-NR-GAN-I}, we apply a Gaussian approximation to the Poisson noise to make it tractable and differentiable. However, through experiments, we found that this approximation causes the performance degradation even using AmbientGAN, which knows all information about the noise. A possible reason is that powerful $G_{\bm x}$ attempts to fill in the discretized gap caused by this approximation. To alleviate the effect, we apply an anti-alias (or low-pass) filter~\cite{RZhangICML2019} to ${\bm x}$ before providing to $D_{\bm y}$. In particular, we found that applying vertical and horizontal blur filters respectively and providing both to $D_{\bm y}$ works well. In the experiments, we apply this technique to all GANs in the Poisson or Poisson-Gaussian noise setting.\footnote{Strictly speaking, this strategy goes against the assumptions of SD-NR-GAN-II and -III because they are agnostic to the signal-noise relationship. However, in this main text, we do that to focus on comparison of the generator performance.} We discuss the effect with and without this technique in Appendix~\ref{subsec:effect_blur}.

\section{Experiments}
\label{sec:experiments}

\subsection{Comprehensive study}
\label{subsec:comprehensive_study}

To advance the research on noise robust image generation, we first conducted a comprehensive study, where we compared various models in diverse noise settings (in which we tested 152 conditions in total).

\smallskip\noindent\textbf{Data setting.}
In this comprehensive study, we used \textsc{CIFAR-10}~\cite{AKrizhevskyTech2009}, which contains $60k$ $32 \times 32$ natural images, partitioned into $50k$ training and $10k$ test images. We selected this dataset because it is commonly used to examine the benchmark performance of generative models (also in the study of AmbientGAN~\cite{ABoraICLR2018}); additionally, the image size is reasonable for a large-scale comparative study. Note that we also conducted experiments using more complex datasets in Section~\ref{subsec:complex}. With regard to noise, we tested 16 noises, shown in Figure~\ref{fig:noise_examples}. See the caption for their details.

\smallskip\noindent\textbf{Compared models.}
In addition to the models in Figures~\ref{fig:models_si} and \ref{fig:models_sd}, we tested several baselines. As comparative GAN models, we examined four models: (1) Standard \textbf{GAN}. (2) \textbf{P-AmbientGAN} (parametric AmbientGAN), a straightforward extension of AmbientGAN, which has a single trainable parameter $\sigma$. As with SI-NR-GAN-I and SD-NR-GAN-I, we construct this model for Gaussian, multiplicative Gaussian, and Poisson noises and generate the noise with $\sigma$ using a reparameterization trick~\cite{DKingmaICLR2014}. (3) \textbf{SI-NR-GAN-0} (Figure~\ref{fig:models_si}(b)), which has the same generators as SI-NR-GANs but has no constraint on $G_{\bm n}$. (4) \textbf{SD-NR-GAN-0}, which has the same generators as SD-NR-GAN-II and -III but has no constraint on $G_{\bm n}$.

We also examined the performance of learning GANs using denoised images (\textbf{denoiser+GANs}). As a denoiser, we investigated four methods. As typical model-based methods, we used (1) \textbf{GAT-BM3D}~\cite{MMakitaloTIP2012} and (2) \textbf{CBM3D}~\cite{KDabovICIP2007} for Poisson/Poisson-Gaussian noise (Figure~\ref{fig:noise_examples}(M)--(P)) and the other noises, respectively. As discriminative learning methods, we used (3) \textbf{N2V} (Noise2Void)~\cite{AKrullCVPR2019} and (4) \textbf{N2N} (Noise2Noise)~\cite{JLehtinenICML2018}. N2V can be used in the same data setting as ours (i.e., only noisy images are available for training), while N2N requires noisy image \textit{pairs} for training. We used N2N because it is commonly used as the upper bound of self-supervised learning methods (e.g., N2V).

\smallskip\noindent\textbf{Evaluation metrics.}
We used the Fr\'{e}chet inception distance (FID)~\cite{MHeuselNIPS2017} as an evaluation metric because its validity has been demonstrated in large-scale studies on GANs~\cite{MLucicNeurIPS2018,KKurachICML2019}, and because the sensitivity to the noise has also been shown~\cite{MHeuselNIPS2017}. The FID measures the distance between real and generative distributions and a smaller value is better.

\smallskip\noindent\textbf{Implementation.}
We implemented GANs using the ResNet architectures~\cite{KHeCVPR2016} and trained them using a non-saturating GAN loss~\cite{IGoodfellowNIPS2014} with a real gradient penalty regularization~\cite{LMeschederICML2018}. In NR-GANs, we used similar architectures for $G_{\bm x}$ and $G_{\bm n}$. As our aim is to construct a general model applicable to various noises, we examined the performance when the training settings are fixed regardless of the noise. We provide the implementation details in Appendix~\ref{subsec:detail_comprehensive_study}.

\smallskip\noindent\textbf{Results on signal-independent noises.}
The upper part of Table~\ref{tab:fid_cifar10} summarizes the results on signal-independent noises. In P-AmbientGAN and SI-NR-GANs, we defined the distribution type as Gaussian for all noise settings and analyzed the effect when the noise is beyond assumption. Our main findings are the following:

\smallskip\noindent\textit{(1) Comparison among GAN models.} As expected, AmbientGAN tends to achieve the best score owing to the advantageous training setting, while the best SI-NR-GAN shows a competitive performance (with a difference of 3.3 in the worst case). P-AmbientGAN is defeated by SI-NR-GAN-I in all cases. These results indicate that our two-generator model is reasonable when training a noise generator and image generator simultaneously.

\smallskip\noindent\textit{(2) Comparison between SI-NR-GANs and denoiser+GANs.} The best SI-NR-GAN outperforms the best denoiser+GAN in most cases (except for (G)). In particular, pixel-wise correlated noises (G)(H) are intractable for denoiser+GANs except for N2N+GAN, which uses additional supervision, while SI-NR-GAN-II works well and outperforms the baseline models by a large margin (with a difference of over 100).

\smallskip\noindent\textit{(3) Comparison among SI-NR-GANs.} SI-NR-GAN-II shows the stable performance across all cases (the difference to the best SI-NR-GAN is within 3.1). SI-NR-GAN-I shows the best or competitive performance in Gaussian (A)--(D) or near Gaussian noise (F); however, the performance degrades when the distribution is beyond assumption (E)(G)(H).

\begin{table}[tb]
  \centering
  \scriptsize{
    \begin{tabularx}{\columnwidth}{cYYYYYYYY} \toprule
      \textbf{Signal-} & (A) & (B) & (C) & (D) & (E) & (F) & (G) & (H)
      \\
      \textbf{independent} & \textbf{AGF} & \textbf{AGV} & \textbf{LGF} & \textbf{LGV} & \textbf{U} & \textbf{MIX} & \textbf{BG} & \textbf{A+G}
      \\ \midrule
      AmbientGAN$^{\dag}$
      & 26.7 & 28.0 & 21.8 & 21.7 & 28.3 & 25.1 & 30.3 & 40.8
      \\
      P-AmbientGAN
      & 33.9 & 122.2 & 38.8 & 38.0 & 43.0 & 32.3 & 164.2 & 269.7
      \\
      GAN
      & 145.8 & 136.0 & 38.8 & 38.8 & 146.4 & 125.6 & 165.3 & 265.9
      \\ \midrule
      SI-NR-GAN-0
      & 40.7 & 39.5 & 23.1 & 24.3 & 38.6 & 32.7 & 71.6 & 139.7
      \\ 
      SI-NR-GAN-I
      & \textbf{26.7} & \textbf{27.5} & \textbf{22.1} & 22.4 & 40.1 & \textbf{24.8} & 163.4  & 253.2
      \\
      SI-NR-GAN-II
      & 29.8 & 29.7 & \textbf{22.1} & \textbf{21.7} & \textbf{31.6} & 26.5 & \textbf{32.2} & \textbf{44.0}
      \\ \midrule
      CBM3D+GAN
      & 35.1 & 38.4 & 37.0 & 33.9 & 38.9 & 30.2 & 136.6 & 169.1
      \\
      N2V+GAN
      & 34.6 & 36.7 & 22.7 & 22.6 & 36.4 & 32.0 & 163.8 & 247.8
      \\ 
      N2N+GAN$^{\ddag}$
      & 33.5 & 36.5 & 22.4 & 22.0 & 32.4 & 30.7 & 29.5 & 48.3
      \\ \bottomrule \toprule
      \textbf{Signal-} & (I) & (J) & (K) & (L) & (M) & (N) & (O) & (P)
      \\
      \textbf{dependent} & \textbf{MGF} & \textbf{MGV} & \textbf{{\tiny A}+I} & \textbf{A+I} & \textbf{PF} & \textbf{PV} & \textbf{{\tiny A}+M} & \textbf{A+M}
      \\ \midrule
      AmbientGAN$^\dag$
      & 21.4 & 21.8 & 21.9 & 27.4 & 31.3 & 32.3 & 30.9 & 35.3
      \\
      P-AmbientGAN
      & 27.1 & 68.7 & 39.7 & 137.7 & 43.8 & 100.7 & 43.0 & 94.2
      \\
      GAN
      & 82.7 & 77.4 & 93.2 & 155.8 & 152.4 & 160.1 & 149.1 & 175.8
      \\ \midrule
      SD-NR-GAN-0
      & 82.7 & 59.5 & 69.9 & 75.1 & 71.7 & 70.2 & 72.0 & 69.0
      \\
      SD-NR-GAN-I
      & \textbf{22.5} & \textbf{23.0} & 25.3 & 112.4 & \textbf{30.8} & \textbf{32.0} & \textbf{31.4} & 70.6
      \\
      SD-NR-GAN-II
      & 24.4 & 24.2 & \textbf{23.3} & \textbf{28.5} & 34.0 & 33.9 & 34.0 & \textbf{35.4}
      \\
      SD-NR-GAN-III
      & 37.5 & 33.4 & 33.5 & 33.9 & 53.1 & 55.1 & 52.4 & 47.2
      \\ \midrule
      CBM3D+GAN
      & 26.9 & 27.8 & 27.6 & 40.0 & -- & -- & -- & --
      \\
      \:\:GAT-BM3D+GAN\:\:
      & -- & -- & -- & -- & 38.4 & 40.2 & 38.7 & 50.1
      \\
      N2V+GAN
      & 25.8 & 26.6 & 26.7 & 36.4 & 37.1 & 38.3 & 37.5 & 41.2
      \\
      N2N+GAN$^\ddag$
      & 24.9 & 26.6 & 26.2 & 34.1 & 36.7 & 39.7 & 36.4 & 39.5
      \\ \bottomrule
    \end{tabularx}
  }
  \caption{
    \textbf{Comparison of FID on \textsc{CIFAR-10}.} A smaller value is better. We compared 152 conditions. The second and thirteenth rows denote the abbreviations defined in Figure~\ref{fig:noise_examples}. We report the median score across three random seeds. The symbol $^{\dag}$ indicates that the ground-truth noise models are given. The symbol $^{\ddag}$ denotes that noisy image pairs are given during the training. The other models are trained using only noisy images (not including pairs) without complete noise information. Bold font indicates the best score except for the models denoted by ${^{\dag\ddag}}$.
  }
  \label{tab:fid_cifar10}
  \vspace{-3.5mm}
\end{table}

\smallskip\noindent\textbf{Results on signal-dependent noises.}
The lower part of Table~\ref{tab:fid_cifar10} lists the results on signal-dependent noises. In P-AmbientGAN and SD-NR-GAN-I, we defined the distribution type as multiplicative Gaussian and Poisson in (I)--(L) and (M)--(P), respectively. With regard to a \textit{comparison among GAN models} and \textit{comparison between SD-NR-GANs and denoiser+GANs}, similar findings (i.e., the best SD-NR-GAN is comparable with AmbientGAN and outperforms the best denoiser+GAN) are observed; therefore, herein we discuss a \textit{comparison among SD-NR-GANs}. SD-NR-GAN-II and -III stability work better than SD-NR-GAN-0. Among the two, SD-NR-GAN-II, which has a stronger assumption, outperforms SD-NR-GAN-III in all cases (with a difference of over 5.4). SD-NR-GAN-I shows the best or competitive performance when noises are within or a little over assumption (I)--(K)(M)--(O); however, when the unexpected noise increases (L)(P), the performance degrades.

\begin{table}[tb]
  \centering
  \scriptsize{
    \begin{tabularx}{\columnwidth}{cYYYY} \toprule
      \:\:\multirow{3}{*}{\textbf{Signal-independent}}\:\: & \multicolumn{3}{c}{\textbf{\textsc{LSUN Bedroom}}} & \textbf{\textsc{FFHQ}}
      \\ \cmidrule(lr){2-4} \cmidrule{5-5}
      & (A) & (B) & (G) & (A)
      \\
      & \textbf{AGF} & \textbf{AGV} & \textbf{BG} & \textbf{AGF}
      \\ \midrule
      AmbientGAN$^{\dag}$
      & 19.4 & 25.0 & 9.7 & 28.3
      \\
      GAN
      & 98.9 & 100.9 & 125.3 & 81.6
      \\ \midrule
      SI-NR-GAN-I
      & \textbf{13.8} & \textbf{14.2} & 128.2 & \textbf{35.7}
      \\
      SI-NR-GAN-II
      & 15.7 & 16.8 & \textbf{10.8} & 37.1
      \\ \bottomrule \toprule
      \multirow{3}{*}{\textbf{Signal-dependent}} & \multicolumn{3}{c}{\textbf{\textsc{LSUN Bedroom}}} & \textbf{\textsc{FFHQ}}
      \\ \cmidrule(lr){2-4} \cmidrule{5-5}
      & (I) & (L) & (M) & (I)
      \\
      & \textbf{MGF} & \textbf{A+I} & \textbf{PF} & \textbf{MGF}
      \\ \midrule
      AmbientGAN$^{\dag}$
      & 11.7 & 19.2 & 32.6 & 18.7
      \\
      GAN
      & 54.0 & 109.8 & 121.7 & 48.0
      \\ \midrule
      SD-NR-GAN-I
      & \textbf{11.6} & 55.4 & \textbf{23.3} & \textbf{26.5}
      \\
      SD-NR-GAN-II
      & 21.7 & \textbf{15.0} & 42.8 & 49.0
      \\
      SD-NR-GAN-III
      & 50.7 & 53.1 & 138.6 & 37.2
      \\ \bottomrule
    \end{tabularx}
  }
  \caption{
    \textbf{Comparison of FID on \textsc{LSUN Bedroom} and \textsc{FFHQ}.} A smaller value is better. Because the training is time-consuming, experiments were run once. The notation is the same as that in Table~\ref{tab:fid_cifar10}.
  }
  \label{tab:fid_complex}
  \vspace{-3.5mm}
\end{table}

\smallskip\noindent\textbf{Summary.}
Through the comprehensive study, we confirm the following: (1) NR-GANs work reasonably well comparing to other GAN models and denoiser+GANs. (2) Weakly constrained NR-GANs stability work well across various settings, while (3) strongly constrained NR-GANs show a better performance when noise is within assumption.\footnote{As further analyses, we examined the effect of diversity-sensitive regularization (Appendix~\ref{subsec:effect_ds_regularization}); the effect of blur filtering (Appendix~\ref{subsec:effect_blur}); performance on the clean dataset (Appendix~\ref{subsec:performance_clean}); performance on the semi-noisy datasets (Appendix~\ref{subsec:performance_semi_noisy}); performance on the mixed datasets (Appendix~\ref{subsec:performance_mixed}); and performance of end-to-end denoiser+GAN (Appendix~\ref{subsec:performance_end2end_denoiserGAN}). Moreover, we conducted an ablation study on transformations in SI-NR-GAN-II (Appendix~\ref{subsec:ablation_transformations}); a comparative study on transformations in SD-NR-GAN-III (Appendix~\ref{subsec:comparative_transformations}); an ablation study on implicit relational learning in SD-NR-GAN-II and -III (Appendix~\ref{subsec:ablation_implicit_learning}); and a generality analysis of SD-NR-GAN-II and -III (Appendix~\ref{subsec:generality_sd_nr_gans}). We also show examples of generated images in Appendix~\ref{subsec:examples_comprehensive_study}. See Appendices for their details.}

\subsection{Evaluation on complex datasets}
\label{subsec:complex}

Inspired by the resent large-scale study on GANs~\cite{KKurachICML2019}, we also examined the performance on more complex datasets. Referring to this study, we used the $128 \times 128$ versions of \textsc{LSUN Bedroom}~\cite{FYuArXiv2015} and \textsc{FFHQ}~\cite{TKarrasCVPR2019}.\footnote{Strictly speaking, the previous study~\cite{KKurachICML2019} used CelebA-HQ~\cite{TKarrasICLR2018} instead of FFHQ. The reason why we used FFHQ is that FFHQ is the latest and more challenging dataset that includes vastly more variation.} \textsc{LSUN Bedroom} contains about 3 million bedroom images, randomly split into training and test sets in the ratio of 99 to 1. \textsc{FFHQ} contains $70k$ face images, partitioned into $60k$ training and $10k$ test images. As these datasets are calculationally demanding, we selected six noises for \textsc{LSUN Bedroom} and two noises for \textsc{FFHQ}. We provide the implementation details in Appendix~\ref{subsec:detail_complex}.

Table~\ref{tab:fid_complex} list the results. Just like the \textsc{CIFAR-10} results, we found that the best NR-GAN outperforms standard GAN and its performance is closer to that of AmbientGAN. In contrast, differently from the CIFAR-10 results, we found that in complex datasets, some weakly constrained SD-NR-GANs suffer from learning difficulty (e.g., SD-NR-GAN-III in \textsc{LSUN Bedroom} (M)). This is undesirable but understandable because in complex datasets it is highly challenging to isolate noise from the dependent signal without an explicit knowledge about their dependency. This is related to GAN training dynamics and addressing this limitation is our future work. As reference, we provide qualitative results in Appendix~\ref{subsec:examples_complex}.

\subsection{Application to image denoising}
\label{subsec:application}

NR-GANs can generate an image and noise, respectively. By utilizing this, we create clean and noisy image pairs synthetically and use them for learning a denoiser. We call this method \textit{GeneratedNoise2GeneratedClean} (\textit{GN2GC}). In particular, we employed the generators that achieve the best FID in Table~\ref{tab:fid_complex} (denoted by bold font).\footnote{We provide other case results in Appendix~\ref{subsec:analysis_gn2gc}.} Note that NR-GANs are trained only using noisy images; therefore, GN2GC can be used in the same data setting as self-supervised learning methods (N2V~\cite{AKrullCVPR2019} and N2S~\cite{JBatsonICML2019}). We used the same training and test sets used in Section~\ref{subsec:complex}. We present the implementation details in Appendix~\ref{subsec:detail_application}.

We summarize the results in Table~\ref{tab:psnr}. We found that GN2GC not only outperforms the state-of-the-art self-supervised learning methods (N2V and N2S) but also is comparable with N2N, which learns in advantageous conditions. The requirement for pre-training GANs could narrow the applications of GN2GC; however, we believe that its potential for image denoising would increase along with rapid progress of GANs. We show examples of denoised images in Appendix~\ref{subsec:examples_application}.

\begin{table}[tb]
  \centering
  \scriptsize{
    \begin{tabularx}{\columnwidth}{cYYYYYYYY} \toprule
      & \multicolumn{3}{c}{\textbf{\textsc{LSUN Bedroom}}} & \textbf{\textsc{FFHQ}} & \multicolumn{3}{c}{\textbf{\textsc{LSUN Bedroom}}} & \textbf{\textsc{FFHQ}}
      \\ \cmidrule(lr){2-4} \cmidrule{5-5} \cmidrule(lr){6-8} \cmidrule{9-9}
      & (A) & (B) & (G) & (A) & (I) & (L) & (M) & (I)
      \\
      & \textbf{AGF} & \textbf{AGV} & \textbf{BG} & \textbf{AGF} & \textbf{MGF} & \textbf{A+I} & \textbf{PF} & \textbf{MGF}
      \\ \midrule
      N2C$^{\sharp}$
      & 32.90 & 33.06 & 29.67 & 31.93 & 36.70 & 32.26 & 31.77 & 36.37
      \\
      N2N$^{\ddag}$
      & 32.30 & 32.23 & 28.76 & 31.33 & 35.99 & 31.36 & 30.55 & 35.88
      \\
      N2V
      & 31.98 & 31.85 & 20.73 & 30.95 & 35.37 & 31.09 & 30.30 & 34.95
      \\
      N2S
      & 31.79 & 31.70 & 20.74 & 30.74 & 35.12 & 30.94 & 30.19 & 34.67
      \\
      GN2GC
      & \textbf{32.36} & \textbf{32.47} & \textbf{26.61} & \textbf{31.34} & \textbf{36.01} & \textbf{31.62} & \textbf{31.08} & \textbf{35.69}
      \\
      CBM3D
      & 31.41 & 31.54 & 20.75 & 30.29 & 33.60 & 30.43 & -- & 32.73
      \\
      \:GAT-BM3D\:
      & -- & -- & -- & -- & -- & -- & 29.80 & --
      \\ \bottomrule
    \end{tabularx}
  }
  \caption{
    \textbf{Comparison of PSNR on \textsc{LSUN Bedroom} and \textsc{FFHQ}.} A larger value is better. We report the median score across three random seeds. The symbols $^{\sharp\ddag}$ indicate that the models are trained in advantageous conditions ($^{\sharp}$clean target images and $^{\ddag}$noisy image pairs are given, respectively). The other models are trained using only noisy images (not including pairs). Bold font indicates the best score except for the models denoted by $^{\sharp\ddag}$.
  }
  \label{tab:psnr}
  \vspace{-3.5mm}
\end{table}

\section{Conclusion}
\label{sec:conclusion}

To achieve noise robust image generation without full knowledge of the noise, we developed a new family of GANs called NR-GANs which learn a noise generator with a clean image generator, while imposing a distribution or transformation constraint on the noise generator. In particular, we introduced five variants: two SI-NR-GANs and three SD-NR-GANs, which have different assumptions. We examined the effectiveness and limitations of NR-GANs on three benchmark datasets and demonstrated the applicability in image denoising. In the future, we hope that our findings facilitate the construction of a generative model in a real-world scenario where only noisy images are available.

\section*{Acknowledgement}

We thank Naoya Fushishita, Takayuki Hara, and Atsuhiro Noguchi for helpful discussions. This work was partially supported by JST CREST Grant Number JPMJCR1403, and partially supported by JSPS KAKENHI Grant Number JP19H01115.  

{\small
\bibliographystyle{ieee_fullname}
\bibliography{refs}
}

\clearpage
\appendix
\section{Further analyses}
\label{sec:further_analyses}

In this section, we provide further analyses mentioned in the main text.

\subsection{Effect of diversity-sensitive regularization}
\label{subsec:effect_ds_regularization}

We first discuss the impact of the diversity-sensitive regularization (DS regularization)~\cite{DYangICLR2019} introduced in Section~\ref{subsec:convergence_speed_difference}. As described, we used this regularization to prevent $G_{\bm n}$ from resulting in a mode collapse. In this section, we first detail DS regularization, followed by the experimental results.

DS regularization is defined as follows:
\begin{flalign}
  \label{eqn:DS_regularizatioon}
  \max_{G_{\bm n}} \lambda \mathbb{E}_{{\bm z}_{\bm n}^1, {\bm z}_{\bm n}^2 \sim p({\bm z}_{\bm n})} \left[ \min \left( \frac{\| G_{\bm n}({\bm z}_{\bm n}^1) - G_{\bm n}({\bm z}_{\bm n}^2) \|} {\| {\bm z}_{\bm n}^1 - {\bm z}_{\bm n}^2 \|}, \tau \right) \right],
\end{flalign}
where $\lambda$ controls the importance of DS regularization, and $\tau$ is a boundary for ensuring numerical stability. Intuitively, when a mode collapse occurs and $G_{\bm n}$ produces a deterministic output, Equation~\ref{eqn:DS_regularizatioon} reaches close to its minimum because $G_{\bm n}({\bm z}_{\bm n}^1) \approx G_{\bm n}({\bm z}_{\bm n}^2)$ for all ${\bm z}_{\bm n}^1, {\bm z}_{\bm n}^2 \sim p({\bm z}_{\bm n})$. To avoid this, we regularize $G_{\bm n}$ to maximize Equation~\ref{eqn:DS_regularizatioon} and promote $G_{\bm n}$ to produce diverse outputs depending on ${\bm z}_{\bm n}$.

During the experiments described in Section~\ref{subsec:comprehensive_study}, we set $\tau$ to zero, following the implementation of a study on DS regularization~\cite{DYangICLR2019}. Furthermore, we fixed $\lambda = 0.02$ and applied DS regularization to all NR-GANs to investigate the performance when the training parameters were set.

To further analyze the effect of DS regularization, herein we examined the performance when $\lambda$ is changed. We show qualitative results in Figure~\ref{fig:examples_cifar10_ds}. As shown, the diversity of the noise increases as $\lambda$ increases (in particular, in Figure~\ref{fig:examples_cifar10_ds}(c)).

We also analyzed quantitatively. In particular, we examined both a case in which the images are noisy and a case in which the images are clean. In the former case, DS regularization will be useful because the diversity of the noise is significant, while in the latter case, DS regularization will be superfluous because there is no diversity in the noise.

Figure~\ref{fig:ds} shows a comparison of FID using a different strength of DS regularization. We found that, for the noisy images (Figure~\ref{fig:noise_examples}(A)(B)(I)(M)), all models improve the FID when using the appropriate strength of DS regularization (in particular, $\lambda \in [0.01, 0.02]$). We also confirm that when using $\lambda \in [0.01, 0.02]$, the negative effect is relatively small (the difference is within 1.0) in the clean dataset, where the diversity in the noise is not required. Based on these results, we argue that the incorporation of appropriate DS regularization is reasonable for the NR-GAN framework. As mentioned in the original study on DS regularization~\cite{DYangICLR2019}, learning $\lambda$ from the training data remains an interesting area for future research.

\begin{figure}[thb!]
  \centering
  \includegraphics[width=\columnwidth]{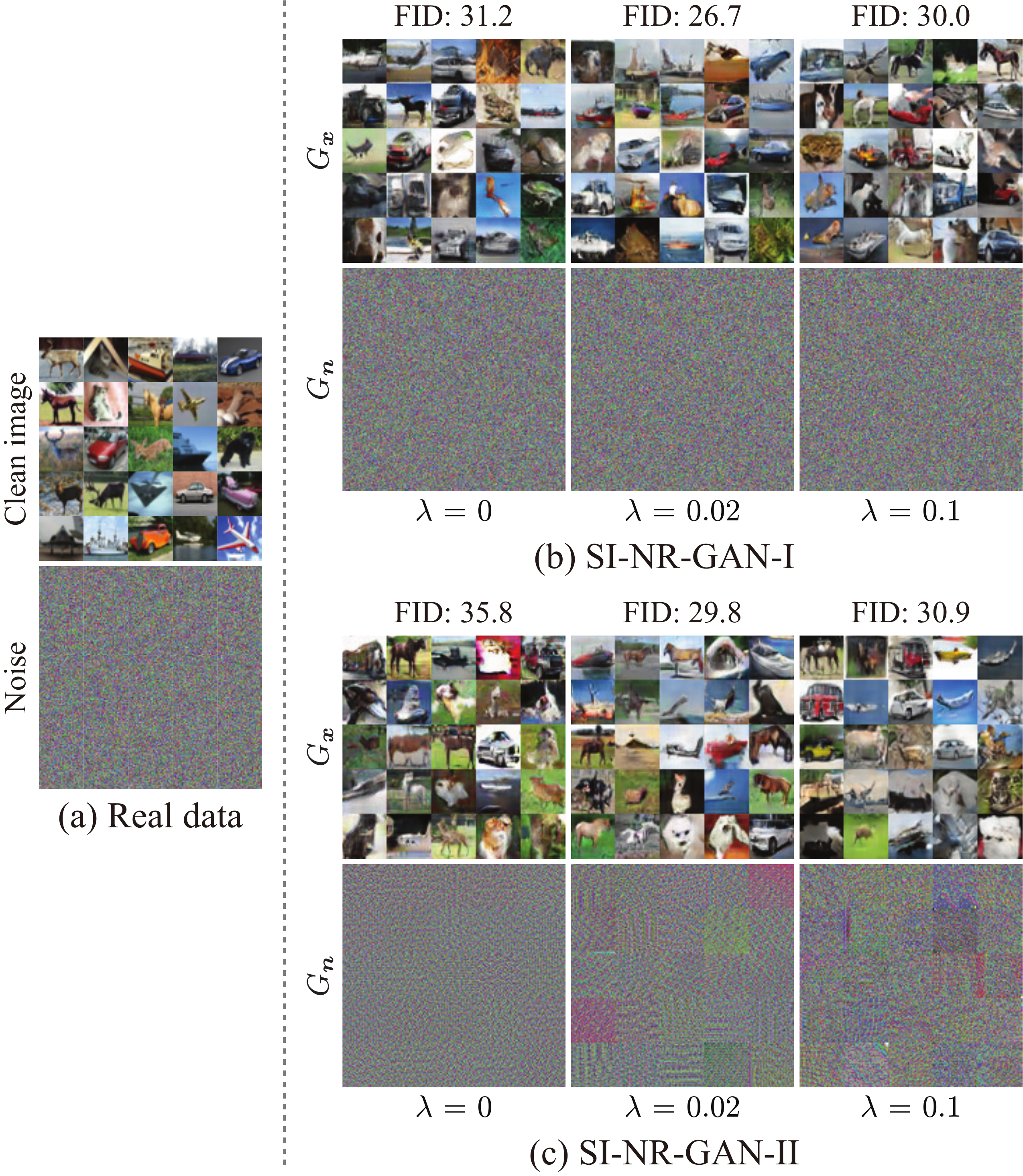}
  \caption{
    \textbf{Examples of images and noises generated when using different strength of diversity-sensitive regularization.} In particular, we show examples for \textsc{CIFAR-10} with additive Gaussian noise (Figure~\ref{fig:noise_examples}(A)). For each case, we show $5 \times 5$ samples. As $\lambda$ increases, the diversity of the noise also increases.
  }
  \label{fig:examples_cifar10_ds}
\end{figure}

\begin{figure}[tb!]
  \centering
  \includegraphics[width=\columnwidth]{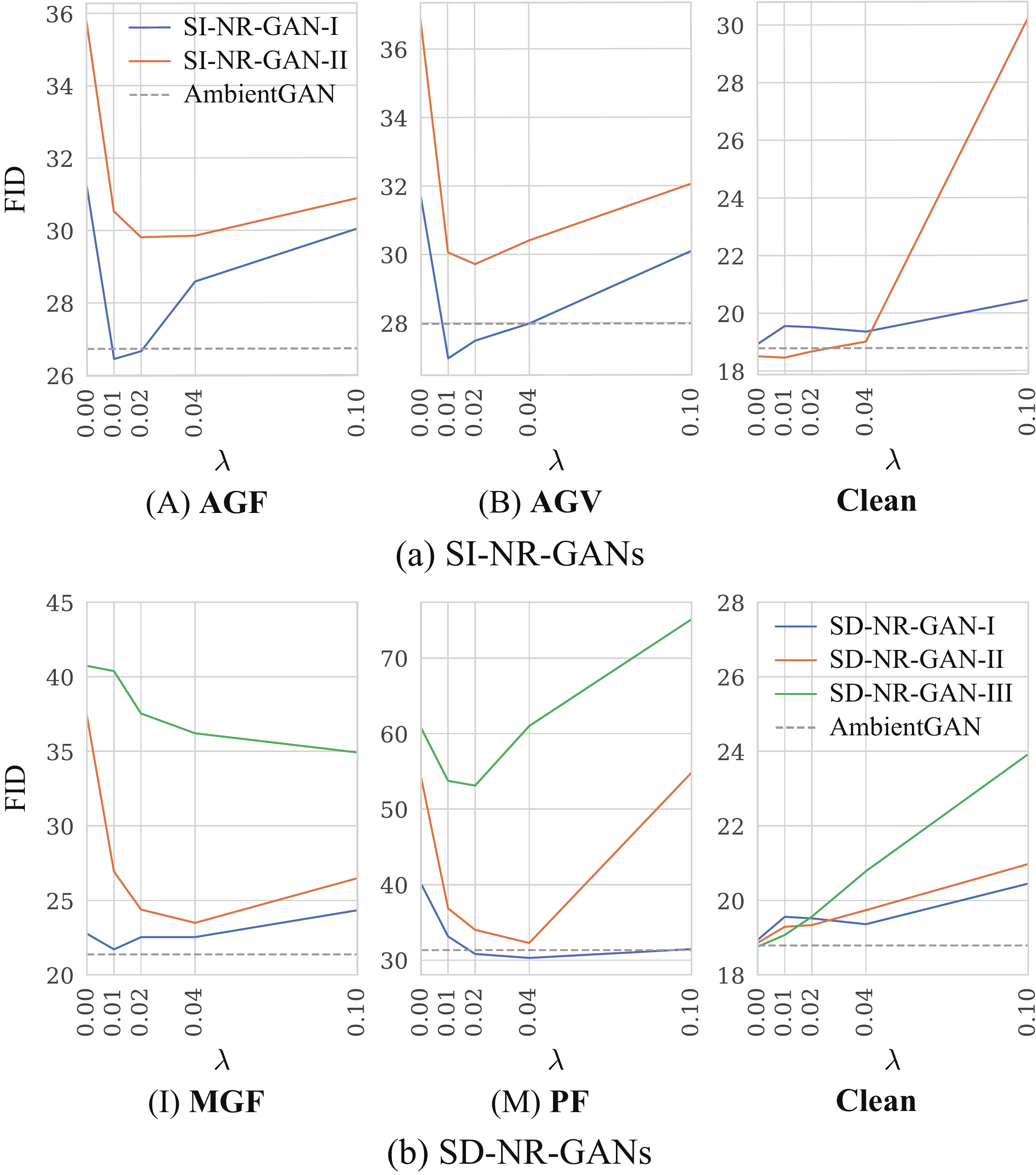}
  \caption{
    \textbf{Comparison of FID on \textsc{CIFAR-10} using different strength of diversity-sensitive regularization.} We report the median score across three random seeds. A smaller value is better.
  }
  \label{fig:ds}
  \vspace{-8mm}
\end{figure}

\begin{figure*}[t!]
  \centering
  \includegraphics[width=0.87\textwidth]{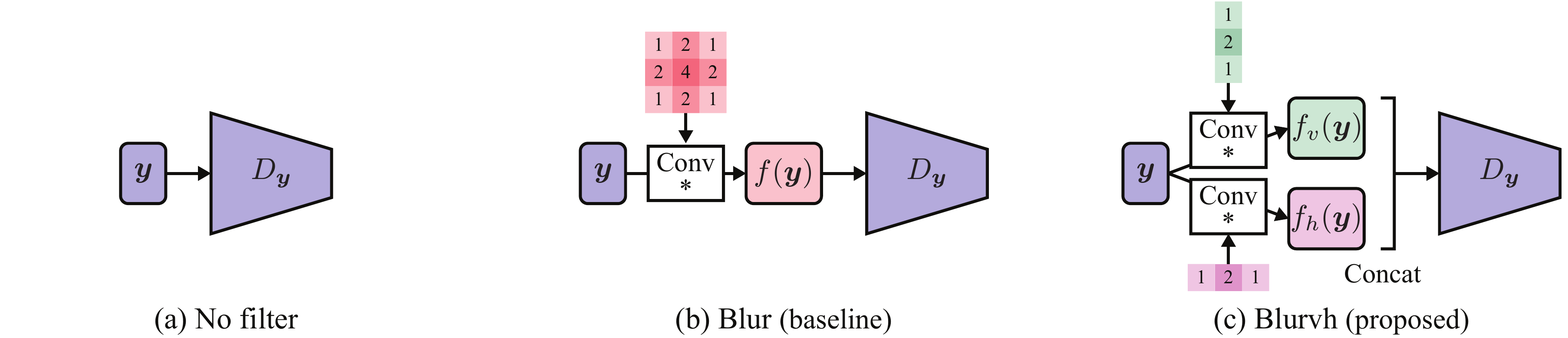}
  \caption{
    \textbf{Comparison of blur filters.} ${\bm y}$ is a real or generated image. In Blurvh (c), we apply both vertical and horizontal blur filters to ${\bm y}$.
  }
  \label{fig:filter}
  \vspace{-2mm}
\end{figure*}

\subsection{Effect of blur filtering}
\label{subsec:effect_blur}

As described in Section~\ref{subsec:approximation_degradation}, we apply blur (i.e., low-pass) filters to Poisson noise to alleviate the difficulty in learning a discrete distribution. In particular, we apply both vertical and horizontal blur filters to an image ${\bm y}$; concatenate the two filtered results; and provide the concatenation to $D_{\bm y}$, as shown in Figure~\ref{fig:filter}(c). We call this approach \textit{Blurvh}. By focusing the filter in one direction, we can alleviate the discretized effect in that direction while preserving the original structure in another direction.

To clarify the effect of \textit{Blurvh}, we compared it with a non-filtered model (\textit{No filter}; Figure~\ref{fig:filter}(a)) and a standard blur filtering model (\textit{Blur}; Figure~\ref{fig:filter}(b)). Table~\ref{tab:fid_blur}(a) lists the results for Poisson noise (Figure~\ref{fig:noise_examples}(M)(N)) and Poisson-Gaussian noise (Figure~\ref{fig:noise_examples}(O)(P)). We observed that, under Poison noise (M)(N) and Poisson-noise dominated Poisson-Gaussian noise (O), the proposed \textit{Blurvh} outperforms \textit{No filter} for all GAN models including AmbientGAN (the difference is more than 3.3 in AmbientGAN and more than 5.3 in SD-NR-GANs). By contrast, standard \textit{Blur} is outperformed by \textit{No filter} in most cases (i.e., 10/12, except for (M)(O) in SD-NR-GAN-III). These results verify the superiority of the proposed \textit{Blurvh}. We also found that under Gaussian-noise dominated Poisson-Gaussian noise (P), \textit{No filter} works well in certain cases (i.e., 2/4, in AmbientGAN and SD-NR-GAN-III). We consider this to be due to the additive Gaussian noise helping alleviating the discretization.

To further understand \textit{Blurvh}, we also examined the performance under non-Poisson noise (namely, a continuously distributed noise). In particular, we conducted a test under multiplicative Gaussian noise (Figure~\ref{fig:noise_examples}(I)(J)) and under additive and multiplicative Gaussian noise (Figure~\ref{fig:noise_examples}(K)(L)). Table~\ref{tab:fid_blur}(b) shows the results. Unsurprisingly, we found that \textit{No filter} performs the best in most cases (i.e., 15/16, except for (L) in SD-NR-GAN-I, where the noise is beyond the assumption of SD-NR-GAN-I). This is because, under a continuously distributed noise, blur filtering is superfluous and results in a loss of the original structure. A remarkable finding here is that the negative effect of \textit{Blurvh} is constantly small compared to that of \textit{Blur}, and the degradation from \textit{No filter} is relatively small (with an average difference of 2.5 when excluding (L) in SD-NR-GAN-I, where SD-NR-GAN-I fails to learn a clean image generator because the noise is beyond assumption). Based on these results, we consider that applying \textit{Blurvh} is a reasonable option when the noise distribution is unknown.

\begin{table}[tb]
  \centering
  \scriptsize{
    \begin{tabularx}{\columnwidth}{ccYYYY} \toprule
      \multirow{2}{*}{\textbf{Model}} & \multirow{2}{*}{\textbf{Filter}} & (M) & (N) & (O) & (P)
      \\
      & & \textbf{PF} & \textbf{PV} & \textbf{{\tiny A}+M} & \textbf{A+M}
      \\ \midrule
      \multirow{3}{*}{AmbientGAN}
      & \:\:No filter\:\: & 34.6 & 36.9 & 34.5 & \textbf{32.2}
      \\
      & Blur & 42.3 & 42.2 & 40.4 & 44.9
      \\
      & Blurvh & \textbf{31.3} & \textbf{32.3} & \textbf{30.9} & 35.3
      \\ \midrule
      \multirow{3}{*}{SD-NR-GAN-I}
      & No filter & 41.1 & 44.6 & 40.3 & 91.4
      \\
      & Blur & 51.7 & 57.4 & 55.5 & 72.0
      \\
      & Blurvh & \textbf{30.8} & \textbf{32.0} & \textbf{31.4} & \textbf{70.6}
      \\ \midrule
      \multirow{3}{*}{SD-NR-GAN-II}
      & No filter & 48.5 & 48.4 & 50.1 & 38.1
      \\
      & Blur & 57.8 & 59.1 & 59.8 & 68.8
      \\
      & Blurvh & \textbf{34.0} & \textbf{33.9} & \textbf{34.0} & \textbf{35.4}
      \\ \midrule
      \multirow{3}{*}{\:\:SD-NR-GAN-III\:\:}
      & No filter & 68.9 & 60.4 & 63.8 & \textbf{41.8}
      \\
      & Blur & 55.1 & 66.0 & 56.5 & 61.5
      \\
      & Blurvh & \textbf{53.1} & \textbf{55.1} & \textbf{52.4} & 47.2
      \\ \bottomrule
      \multicolumn{6}{c}{\multirow{2}{*}{\footnotesize{(a) Poisson/Poisson-Gaussian noise}}}
      \\
      \multicolumn{6}{c}{}
      \\ \toprule
      \multirow{2}{*}{\textbf{Model}} & \multirow{2}{*}{\textbf{Filter}} & (I) & (J) & (K) & (L)
      \\
      & & \textbf{MGF} & \textbf{MGV} & \textbf{{\tiny A}+I} & \textbf{A+I}
      \\ \midrule
      \multirow{3}{*}{AmbientGAN}
      & \:\:No filter\:\: & \textbf{21.4} & \textbf{21.8} & \textbf{21.9} & \textbf{27.4}
      \\
      & Blur & 34.3 & 34.7 & 35.5 & 41.7
      \\
      & Blurvh & 25.1 & 25.4 & 25.3 & 31.6
      \\ \midrule
      \multirow{3}{*}{SD-NR-GAN-I}
      & No filter & \textbf{22.5} & \textbf{23.0} & \textbf{25.3} & 112.4
      \\
      & Blur & 36.6 & 36.4 & 36.0 & \textbf{65.0}
      \\
      & Blurvh & 24.6 & 25.4 & 25.4 & 101.5
      \\ \midrule
      \multirow{3}{*}{SD-NR-GAN-II}
      & No filter & \textbf{24.4} & \textbf{24.2} & \textbf{23.3} & \textbf{28.5}
      \\
      & Blur & 37.1 & 36.6 & 37.0 & 53.3
      \\
      & Blurvh & 25.3 & 26.2 & 25.5 & 30.1
      \\ \midrule
      \multirow{3}{*}{\:\:SD-NR-GAN-III\:\:}
      & No filter & \textbf{37.5} & \textbf{33.4} & \textbf{33.5} & \textbf{33.9}
      \\
      & Blur & 48.8 & 51.2 & 48.9 & 45.7
      \\
      & Blurvh & 38.3 & 38.1 & 35.0 & 38.2
      \\ \bottomrule
      \multicolumn{6}{c}{\multirow{2}{*}{\footnotesize{(b) Multiplicative Gaussian/additive and multiplicative Gaussian noise}}}
    \end{tabularx}
  }
  \vspace{1mm}
  \caption{
    \textbf{Comparison of FID on \textsc{CIFAR-10} using different blur filters.} A smaller value is better. The median score for three random seeds is provided. Bold font shows the best score for each model.
  }
  \label{tab:fid_blur}
  \vspace{-2mm}
\end{table}

\subsection{Performance on clean dataset}
\label{subsec:performance_clean}

In the main text, we provide the results on noisy datasets. An interesting question is the performance on a clean dataset. Ideally, it is expected that $G_{\bm n}$ learns \textit{no noise} and outputs a value of zero in this case. To examine this, we conducted experiments on the original (clean) \textsc{CIFAR-10}.

Table~\ref{tab:fid_clean} lists the results. We found that the performance is almost the same (with a difference of within 1.3) except for SD-NR-GAN-0. To examine the reason for the degradation in SD-NR-GAN-0, we show examples of generated images in Figure~\ref{fig:examples_cifar10_clean}. From this figure, we can see that for SD-NR-GAN-0, both $G_{\bm x}$ and $G_{\bm n}$ attempt to generate an image. By contrast, in SD-NR-GAN-II, $G_{\bm x}$ generates an image and $G_{\bm n}$ outputs a value of zero. A similar tendency as with SD-NR-GAN-II is observed in the other NR-GANs. These results verify that our proposed distribution or transformation constraint is useful for making $G_{\bm n}$ learn \textit{no noise}.

\begin{table}[tb]
  \centering
  \scriptsize{
    \begin{tabularx}{0.4\columnwidth}{cY} \toprule
      \multirow{1}{*}{\textbf{Model}}
      & \textbf{Clean}
      \\ \midrule
      GAN & 18.8
      \\
      P-AmbientGAN
      & 18.3
      \\ \midrule
      SI-NR-GAN-0
      & 19.1
      \\
      SI-NR-GAN-I
      & 19.5
      \\      
      SI-NR-GAN-II
      & 18.7
      \\ \midrule
      SD-NR-GAN-0
      & \textit{57.0}
      \\
      SD-NR-GAN-II
      & 19.3
      \\
      \:\:SD-NR-GAN-III\:\:
      & 19.6
      \\ \bottomrule
    \end{tabularx}
  }
  \vspace{1mm}
  \caption{
    \textbf{Comparison of FID on clean \textsc{CIFAR-10}.} A smaller value is better. We omit SD-NR-GAN-I because it is equal to SI-NR-GAN-I under the assumption that it knows the signal-noise relationship (i.e., no relationship here). We give the median score of three random seeds. Italic font indicates the worst score.
  }
  \label{tab:fid_clean}
\end{table}

\begin{figure}[tb]
  \centering
  \includegraphics[width=\columnwidth]{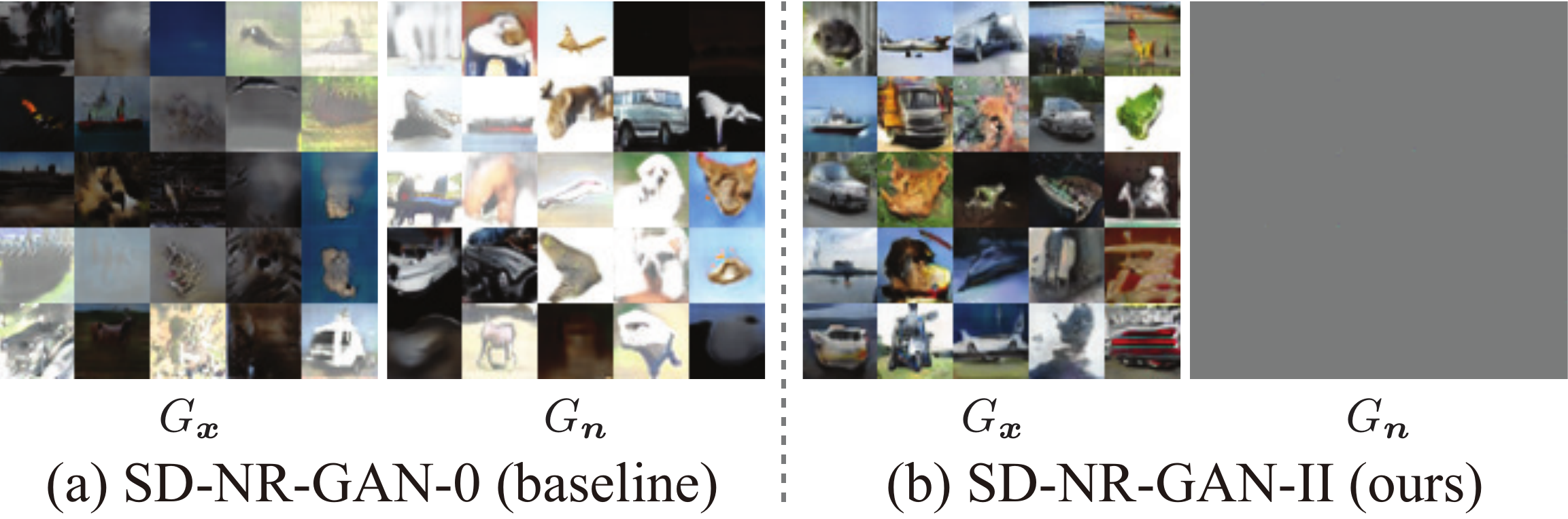}
  \caption{
    \textbf{Examples of images generated for clean \textsc{CIFAR-10}.} We show $5 \times 5$ samples for each case. (a) In SD-NR-GAN-0, which does not regularize $G_{\bm n}$, both $G_{\bm x}$ and $G_{\bm n}$ attempt to generate an image. (b) By contrast, in SD-NR-GAN-II, which imposes a distribution constraint on $G_{\bm n}$, $G_{\bm x}$ generates an image and $G_{\bm n}$ outputs a value of zero (i.e., learns \textit{no noise}).
  }
  \label{fig:examples_cifar10_clean}
  \vspace{-3mm}
\end{figure}

\subsection{Performance on semi-noisy datasets}
\label{subsec:performance_semi_noisy}

In the main text, we conducted experiments on datasets in which all images are noisy. Besides, a practically important question is the performance on a semi-noisy dataset in which parts of images are noisy while the remaining are clean. To answer this question, we conducted experiments on the semi-noisy datasets. In particular, we tested two signal-independent noises, additive Gaussian noise (Figure~\ref{fig:noise_examples}(A)) and Brown Gaussian noise (Figure~\ref{fig:noise_examples}(G)), and two signal-dependent noises, multiplicative Gaussian noise (Figure~\ref{fig:noise_examples}(I)) and additive and multiplicative Gaussian noise (Figure~\ref{fig:noise_examples}(L)). We selected the noise rate $r$ from $r \in \{ 0.25, 0.5, 0.75, 1 \}$.

Table~\ref{tab:fid_semi_noisy} summarizes the results. We found a similar tendency as that described in Section~\ref{subsec:comprehensive_study}. (1) The best NR-GAN is comparable with AmbientGAN. Note that in this case, AmbientGAN requries additional knowledge regarding the rate of noisy samples, whereas NR-GANs can obtain this knowledge in a data-driven manner. (2) The best NR-GAN shows better or competitive performance compared to the best denoiser+GAN in most cases (except for (G) with $r = 0.75$ and $r = 1$, where N2N+GAN, trained under advantageous conditions, performs the best). (3) Weakly constrained NR-GANs (i.e., SI-NR-GAN-II, SD-NR-GAN-II, and SD-NR-GAN-III) stability work well across diverse settings, while strongly constrained NR-GANs (i.e., SI-NR-GAN-I and SD-NR-GAN-I) show a better performance when noise is within assumption (i.e., (A) and (I), respectively) in most cases (except for (A) with $r = 0.25$ and $r = 0.5$, where SI-NR-GAN-II works slightly better than SI-NR-GAN-I). These results validate the usefulness of NR-GANs on semi-noisy datasets.

\begin{table}[tb]
  \centering
  \scriptsize{
    \begin{tabularx}{\columnwidth}{cYYYYYYYY} \toprule
      \multirow{3}{*}{\textbf{Model}}
      & (A) & (A) & (A) & (A) & (G) & (G) & (G) & (G)
      \\
      & \textbf{AGF} & \textbf{AGF} & \textbf{AGF} & \textbf{AGF} & \textbf{BG} & \textbf{BG} & \textbf{BG} & \textbf{BG}
      \\ \cmidrule(l){2-5} \cmidrule(l){6-9}
      & $0.25$ & $0.5$ & $0.75$ & $1$ & $0.25$ & $0.5$ & $0.75$ & $1$
      \\ \midrule
      AmbientGAN$^\dag$
      & 20.7 & 22.2 & 24.1 & 26.7 & 21.4 & 23.3 & 26.4 & 30.3
      \\
      GAN
      & 30.5 & 55.9 & 90.3 &145.8 & 35.9 & 63.7 &106.0 &165.3
      \\ \midrule
      SI-NR-GAN-I
      & 21.5 & 23.9 & \textbf{25.9} & \textbf{26.7} & 36.3 & 65.2 &105.2 &163.4
      \\
      SI-NR-GAN-II
      & \textbf{20.4} & \textbf{23.5} & 27.2 & 29.8 & \textbf{20.8} & \textbf{23.2} & \textbf{27.9} & \textbf{32.2}
      \\ \midrule
      N2V+GAN
      & 21.4 & 24.9 & 29.2 & 34.6 & 39.0 & 65.8 &105.0 &163.8
      \\
      N2N+GAN$^\ddag$
      & 22.3 & 25.0 & 29.0 & 33.5 & 21.0 & 23.6 & 26.0 & 29.5
      \\ \bottomrule
      \multicolumn{9}{c}{\multirow{2}{*}{\footnotesize{(a) Signal-independent noise}}}
      \\
      \multicolumn{9}{c}{}
      \\ \toprule
      \multirow{3}{*}{\textbf{Model}}
      & (I) & (I) & (I) & (I) & (L) & (L) & (L) & (L)
      \\
      & \textbf{MGF} & \textbf{MGF} & \textbf{MGF} & \textbf{MGF} & \textbf{A+I} & \textbf{A+I} & \textbf{A+I} & \textbf{A+I}
      \\ \cmidrule(l){2-5} \cmidrule(l){6-9}
      & $0.25$ & $0.5$ & $0.75$ & $1$ & $0.25$ & $0.5$ & $0.75$ & $1$
      \\ \midrule
      AmbientGAN$^\dag$
      & 19.7 & 20.8 & 21.2 & 21.4 & 21.3 & 23.6 & 25.0 & 27.4
      \\
      GAN
      & 23.4 & 36.0 & 56.4 & 82.7 & 32.6 & 58.2 & 95.5 &155.8
      \\ \midrule
      SD-NR-GAN-I
      & \textbf{20.4} & \textbf{20.6} & \textbf{21.8} & \textbf{22.5} & 33.4 & 60.4 & 93.9 &112.4
      \\
      SD-NR-GAN-II
      & 21.0 & 22.6 & 23.8 & 24.4 & \textbf{20.9} & \textbf{23.4} & \textbf{27.3} & \textbf{28.5}
      \\
      \:\: SD-NR-GAN-III \:\:
      & 21.4 & 23.0 & 27.4 & 37.5 & 22.4 & 25.6 & 30.7 & 33.9
      \\ \midrule
      N2V+GAN
      & 20.5 & 21.8 & 23.1 & 25.8 & 21.6 & 25.7 & 29.9 & 36.4
      \\
      N2N+GAN$^\ddag$
      & 20.3 & 21.6 & 23.4 & 24.9 & 21.5 & 25.6 & 30.5 & 34.1
      \\ \bottomrule
      \multicolumn{9}{c}{\multirow{2}{*}{\footnotesize{(b) Signal-dependent noise}}}
      \\
    \end{tabularx}
  }
  \vspace{1mm}
  \caption{
    \textbf{Comparison of FID on semi-noisy datasets.} A smaller value is better. The third row denotes the noise rate $r$, i.e., what percentage of images include noise. We report the median score across three random seeds. The symbol $^{\dag}$ indicates that the ground-truth noise models are given. The symbol $^{\ddag}$ denotes that noisy image pairs are given during the training. The other models are trained using only noisy images (not including pairs) without complete noise information. Bold font indicates the best score except for the models denoted by ${^{\dag\ddag}}$.
  }
  \label{tab:fid_semi_noisy}
  \vspace{-3mm}
\end{table}

\begin{table}[tb]
  \centering
  \scriptsize{
    \begin{tabularx}{\columnwidth}{cYYYYY} \toprule
      \multirow{3}{*}{\textbf{Model}}
      & (A) & (A)/(E) & (A)/(E) & (A)/(E) & (E)
      \\
      & \textbf{AGF} & \textbf{AGF}/\textbf{U} & \textbf{AGF}/\textbf{U} & \textbf{AGF}/\textbf{U} & \textbf{U}
      \\ \cmidrule(l){2-6}
      & $0$ & $0.25$ & $0.5$ & $0.75$ & $1$
      \\ \midrule
      AmbientGAN$^\dag$
      & 26.7 & 26.9 & 27.5 & 28.3 & 28.3
      \\
      \:\:P-AmbientGAN\:\:
      & 33.9 & 39.9 & 49.6 & 40.5 & 43.0
      \\
      GAN
      & 145.8 & 148.4 & 151.4 & 146.0 & 146.4
      \\ \midrule
      SI-NR-GAN-0
      & 40.7 & 41.1 & 43.9 & 41.0 & 38.6
      \\
      SI-NR-GAN-I
      & \textbf{26.7} & \textbf{27.4} & \textbf{27.2} & 32.1 & 40.1
      \\
      SI-NR-GAN-II
      & 29.8 & 30.5 & 30.8 & \textbf{31.2} & \textbf{31.6}
      \\ \midrule
      CBM3D+GAN
      & 35.1 & 35.9 & 37.1 & 37.2 & 38.9
      \\
      N2V+GAN
      & 34.6 & 36.6 & 36.3 & 36.7 & 36.4
      \\
      N2N+GAN$^\ddag$
      & 33.5 & 35.4 & 33.4 & 33.4 & 32.4
      \\ \bottomrule
      \multicolumn{6}{c}{\multirow{2}{*}{\footnotesize{(a) Mixture of additive Gaussian noise and uniform noise}}}
      \\
      \multicolumn{6}{c}{}
      \\ \toprule
      \multirow{3}{*}{\textbf{Model}}
      & (A) & (A)/(G) & (A)/(G) & (A)/(G) & (G)
      \\
      & \textbf{AGF} & \textbf{AGF}/\textbf{BG} & \textbf{AGF}/\textbf{BG} & \textbf{AGF}/\textbf{BG} & \textbf{BG}
      \\ \cmidrule(l){2-6}
      & $0$ & $0.25$ & $0.5$ & $0.75$ & $1$
      \\ \midrule
      AmbientGAN$^\dag$
      & 26.7 & 27.5 & 29.4 & 29.9 & 30.3
      \\
      \:\:P-AmbientGAN\:\:
      & 33.9 & 130.9 & 132.7 & 139.5 & 164.2
      \\
      GAN
      & 145.8 & 131.9 & 131.1 & 137.7 & 165.3
      \\ \midrule
      SI-NR-GAN-0
      & 40.7 & 33.3 & 36.9 & 44.3 & 71.6
      \\
      SI-NR-GAN-I
      & \textbf{26.7} & 106.3 & 173.3 & 138.0 & 163.4
      \\
      SI-NR-GAN-II
      & 29.8 & \textbf{32.1} & \textbf{31.8} & \textbf{33.3} & \textbf{32.2}
      \\ \midrule
      CBM3D+GAN
      & 35.1 & 46.9 & 70.3 & 99.8 & 136.6
      \\
      N2V+GAN
      & 34.6 & 49.5 & 74.7 & 110.1 & 163.8
      \\
      N2N+GAN$^\ddag$
      & 33.5 & 33.7 & 32.6 & 31.0 & 29.5
      \\ \bottomrule
      \multicolumn{6}{c}{\multirow{2}{*}{\footnotesize{(b) Mixture of additive Gaussian noise and Brown Gaussian noise}}}
    \end{tabularx}
  }
  \vspace{1mm}
  \caption{
    \textbf{Comparison of FID on mixed datasets.} A smaller value is better. The third row presents the mixture rate $\mu$, i.e., what percentage of images contain uniform noise (E) or Brown Gaussian noise (G) is inserted. We report the median score across three random seeds. The notation is the same as that in Table~\ref{tab:fid_semi_noisy}.
  }
  \label{tab:fid_mix}
  \vspace{-3mm}
\end{table}

\subsection{Performance on mixed datasets}
\label{subsec:performance_mixed}

In the main text, we conducted experiments on the datasets in which we assume that the noise parameters, excluding the amount of noise, are fixed across all training images. An interesting question is the performance on the mixed datasets, i.e., partial images containing a certain type of noise and the remaining images including a different type of noise. To investigate this, we conducted experiments on the mixed datasets. In particular, we tested two cases: (1) \textbf{AGF/U}, where partial images contain additive Gaussian noise (Figure~\ref{fig:noise_examples}(A)) while the remaining include uniform noise (Figure~\ref{fig:noise_examples}(E)), and (2) \textbf{AGF/BG}, in which partial images contain additive Gaussian noise (Figure~\ref{fig:noise_examples}(A)) while the remaining include Brown Gaussian noise (Figure~\ref{fig:noise_examples}(G)). We selected the mixture rate $\mu$ from $\mu \in \{ 0, 0.25, 0.5, 0.75, 1 \}$.

Table~\ref{tab:fid_mix} summarizes the results. With regard to a \textit{comparison among GAN models} and \textit{comparison between SI-NR-GANs and denoiser+GANs}, we observed a similar tendency as that described in Section~\ref{subsec:comprehensive_study}, i.e., the best SI-NR-GAN is comparable with AmbientGAN and outperforms the best denoiser+GAN in most cases (except for (A)/(G) with $\mu = 0.75$ and (G), where N2N+GAN, trained under advantageous conditions, performs the best). Hence, herein we discuss a \textit{comparison among SI-NR-GANs} in greater detail. We found that in both mixed datasets, SI-NR-GAN-II, which can be applied to all noises, shows a stable performance (with a difference of within 3.5). By contrast, SI-NR-GAN-I, which assumes Gaussian noise, degrades the performance as the rate of unexpected noise (i.e., uniform or Brown Gaussian noise) increases. In particular, we observed that the mixing of Brown Gaussian noise has a larger impact. The reason for this is that Brown Gaussian noise has a correlation between pixels, which is more difficult to approximate by a Gaussian distribution, compared to uniform noise which is pixel-wise independent. These results indicate the importance of a model selection based on the noise type that may be contained.

\begin{figure}[tb]
  \centering
  \includegraphics[width=\columnwidth]{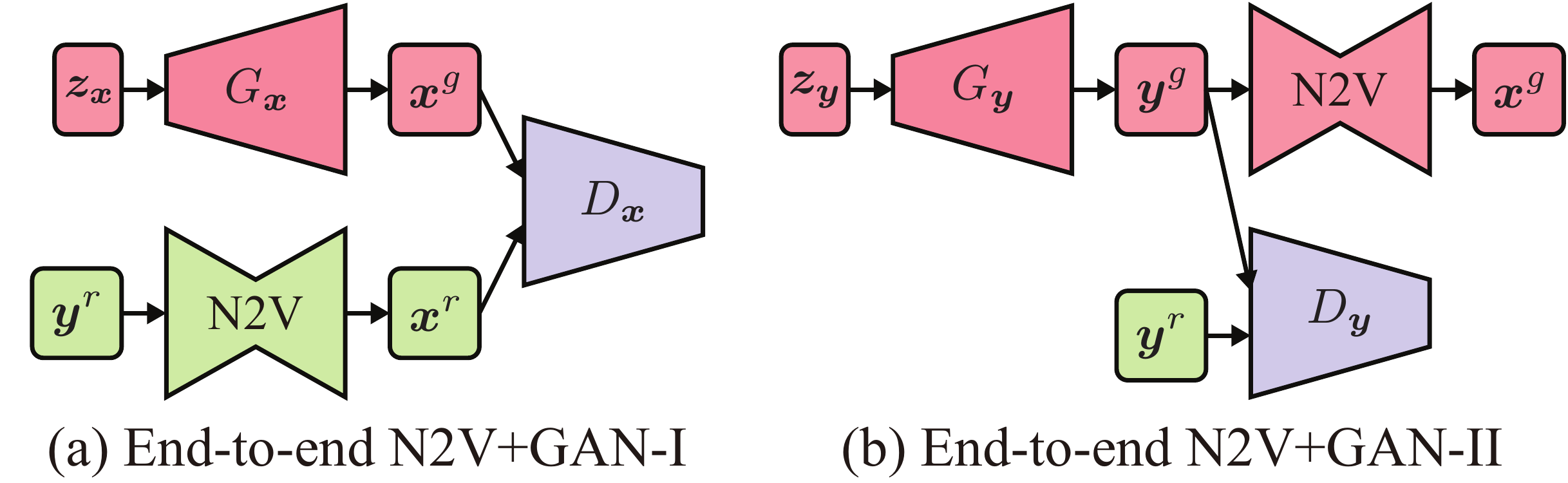}
  \caption{
    \textbf{Architectures of end-to-end denoiser+GANs.} (a) A denoiser (N2V) is trained in a self-supervised manner using a \textit{real} noisy image ${\bm y}^r$. Then, a generator $G_{\bm x}$ is optimized for a \textit{clean} image discriminator $D_{\bm x}$ that distinguishes the generated \textit{clean} image ${\bm x}^g$ from the real \textit{denoised} image ${\bm x}^r$. (b) A generator $G_{\bm y}$ is optimized for a \textit{noisy} image discriminator $D_{\bm y}$ that discriminates the generated \textit{noisy} image ${\bm y}^g$ from the real \textit{noisy} image ${\bm y}^r$. Subsequently, a denoiser (N2V) is trained in a self-supervised manner using the \textit{generated} noisy image ${\bm y}^g$.
  }
  \label{fig:end2end_n2vgan}
  \vspace{-1mm}
\end{figure}

\begin{table}[tb]
  \centering
  \scriptsize{
    \begin{tabularx}{0.5\columnwidth}{cY} \toprule
      \multirow{2}{*}{\textbf{Model}} & (A)
      \\
      & \textbf{AGF}
      \\ \midrule
      N2N+GAN
      & \textbf{34.6}
      \\ \midrule
      End-to-end N2N+GAN-I
      & 37.2
      \\
      \:\: End-to-end N2N+GAN-II \:\:
      & 101.8
      \\ \bottomrule
    \end{tabularx}
  }
  \vspace{1mm}
  \caption{
    \textbf{Comparison of FID on \textsc{CIFAR-10} with additive Gaussian noise (Figure~\ref{fig:noise_examples}(A)) using differently trained N2V+GANs.} A smaller value is better. We give the median score of three random seeds. Bold font indicates the best score.
  }
  \label{tab:fid_end2end_denoiserGAN}
  \vspace{-3mm}
\end{table}

\subsection{Performance of end-to-end denoiser+GAN}
\label{subsec:performance_end2end_denoiserGAN}

In the main text, we used a pretrained denoiser when implementing N2V+GAN and N2N+GAN because it performs better than the end-to-end formulation.

Specifically, we considered two types of end-to-end denoiser+GAN schemes (Figure~\ref{fig:end2end_n2vgan}). Here, for ease of explanation, we focus on the case when N2V~\cite{AKrullCVPR2019} is used as a denoiser. In the main text, we used the same network architectures presented in Figure~\ref{fig:end2end_n2vgan}(a), but adopted a pre-trained denoiser (N2V) because there is no merit in training a denoiser from scratch jointly with GAN because the denoiser is optimized using only \textit{real} noisy images. In this case, we can obtain better denoised images from the start of GAN training by adopting a pre-trained denoiser. As empirical evidence, we show the comparison of FID on \textsc{CIFAR-10} with additive Gaussian noise (Figure~\ref{fig:noise_examples}(A)) in Table~\ref{tab:fid_end2end_denoiserGAN}.

In our preliminary experiments, we also tested the scheme presented in Figure~\ref{fig:end2end_n2vgan}(b). However, we found that it yields poor performance, as listed in Table~\ref{tab:fid_end2end_denoiserGAN}. We consider that this is because in this scheme, the denoiser (N2V) is optimized using \textit{generated} noisy images that do not necessarily satisfy the assumption of N2V (i.e., noise is pixel-wise independent) until the generator is able to recreate a noisy image completely.

\begin{table}[tb]
  \centering
  \scriptsize{
    \begin{tabularx}{\columnwidth}{cYYYYYYYY} \toprule
      \multirow{2}{*}{\textbf{Model}} & (A) & (B) & (C) & (D) & (E) & (F) & (G) & (H)
      \\
      & \textbf{AGF} & \textbf{AGV} & \textbf{LGF} & \textbf{LGV} & \textbf{U} & \textbf{Mix} & \textbf{BG} & \textbf{A+G}
      \\ \midrule
      \multicolumn{1}{l}{\:\:\ \ \ SI-NR-GAN-0}
      & 40.7 & 39.5 & 23.1 & 24.3 & 38.6 & 32.7 & 71.6 & 139.7
      \\
      \multicolumn{1}{l}{\:\:+ Rotation}
      & \textbf{28.3} & 29.8 & 24.0 & 23.9 & 32.2 & 27.6 & 36.1 & 47.7
      \\
      \multicolumn{1}{l}{\:\:+ Shuffle}
      & 30.2 & 30.4 & 24.1 & 24.1 & 32.2 & 27.2 & 36.1 & 47.1
      \\
      \multicolumn{1}{l}{\:\:+ Inversion}
      & 29.5 & 30.3 & 22.3 & 22.0 & 32.1 & 26.8 & 33.6 & 44.3
      \\
      \multicolumn{1}{l}{\:\:+ Rotation \& shuffle}
      & 29.6 & 30.2 & 23.1 & 23.3 & 32.1 & \textbf{26.1} & 36.2 & 46.1
      \\
      \multicolumn{1}{l}{\:\:+ Rotation \& inversion}
      & 29.8 & 30.1 & \textbf{22.1} & 21.9 & 32.3 & 26.8 & 33.4 & \textbf{44.0}
      \\
      \multicolumn{1}{l}{\:\:+ Shuffle \& inversion}
      & 30.5 & 31.0 & 22.2 & 22.6 & 32.6 & 26.7 & 33.0 & 44.4
      \\
      \multicolumn{1}{l}{\:\:+ All (SI-NR-GAN-II)\:\:}
      & 29.8 & \textbf{29.7} & \textbf{22.1} & \textbf{21.7} & \textbf{31.6} & 26.5 & \textbf{32.2} & \textbf{44.0}
      \\ \bottomrule
    \end{tabularx}
  }
  \caption{
    \textbf{Comparison of FID on \textsc{CIFAR-10} with signal-independent noises using different transformations.} A smaller value is better. SI-NR-GAN-0 with all transformations (the last row) is used as \textit{SI-NR-GAN-II} in other experiments. The median score across three random seeds is given. Bold font indicates the best score for each noise.
  }
  \label{tab:ablation_transformations}
  \vspace{-0mm}
\end{table}

\subsection{Ablation study on transformations in SI-NR-GAN-II}
\label{subsec:ablation_transformations}

As described in Section~\ref{subsec:SI-NR-GAN-II}, we use three transformations, i.e., \textit{rotation}, \textit{channel shuffle}, and \textit{color inversion}, in SI-NR-GAN-II. We conducted an ablation study to reveal the impact of the transformations.

Table~\ref{tab:ablation_transformations} summarizes the results. We observed that, in most cases (6/8), the score becomes the best when using \textit{all transformations}. When focusing on the individual transformation, we found that a dataset dependency occurs: under additive Gaussian noise (A)(B), \textit{rotation} is the most useful, whereas under local Gaussian noise (C)(D) and pixel-wise correlated noise (G)(H), \textit{color inversion} is the most effective. Finding the best transformation from the training data would be an interesting area for future study. Refer to Appendix~\ref{subsec:comparative_transformations}, where we discuss the effect of the transformations in SD-NR-GAN-III.

\begin{table}[tb]
  \centering
  \scriptsize{
    \begin{tabularx}{\columnwidth}{cYYYYYYYY} \toprule
      \multirow{2}{*}{\textbf{Model}} & (I) & (J) & (K) & (L) & (M) & (N) & (O) & (P)
      \\
      & \textbf{MGF} & \textbf{MGV} & \textbf{{\tiny A}+I} & \textbf{A+I} & \textbf{PF} & \textbf{PV} & \textbf{{\tiny A}+M} & \textbf{A+M}
      \\ \midrule
      \multicolumn{1}{l}{\:\ \ \ SD-NR-GAN-0}
      & 82.7 & 59.5 & 69.9 & 75.1 & 71.7 & 70.2 & 72.0 & 69.0
      \\
      \multicolumn{1}{l}{\:+ Rotation}
      & 91.4 & 82.5 &102.0 & 75.4 & 88.9 & 82.5 & 86.3 & 63.2
      \\
      \multicolumn{1}{l}{\:+ Shuffle}
      &100.8 & 92.6 & 94.1 & 76.9 & 81.3 & 76.7 & 78.1 & 81.0
      \\
      \multicolumn{1}{l}{\:+ Inversion (SD-NR-GAN-III) \:}
      & \textbf{37.5} & \textbf{33.4} & \textbf{33.5} & \textbf{33.9} & \textbf{53.1} & \textbf{55.1} & \textbf{52.4} & \textbf{47.2}
      \\
      \multicolumn{1}{l}{\:+ All}
      & 63.7 & 58.4 & 57.6 & 38.5 & 82.2 & 83.1 & 81.3 & 70.1
      \\ \bottomrule
    \end{tabularx}
  }
  \caption{
    \textbf{Comparison of FID on \textsc{CIFAR-10} with signal-dependent noises using different transformations.} A smaller value is better. SD-NR-GAN-0 with \textit{color inversion} (the sixth row) is used as \textit{SD-NR-GAN-III} in other experiments.  We report the median score for three random seeds. Bold font indicates the best score for each noise.
  }
  \label{tab:comparative_transformations}
  \vspace{-2mm}
\end{table}

\subsection{Comparative study on transformations in SD-NR-GAN-III}
\label{subsec:comparative_transformations}

As discussed in Section~\ref{subsec:SD-NR-GAN-III}, we only use \textit{color inversion} in SD-NR-GAN-III because \textit{rotation} and \textit{channel shuffle} collapse the per-pixel signal-noise dependency that is included in typical signal-dependent noise (e.g., Figure~\ref{fig:noise_examples}(I)--(P)). For further clarification, we conducted a comparative study on the transformations in SD-NR-GAN-III.

Table~\ref{tab:comparative_transformations} lists the results. From these results, we found that SD-NR-GAN-0 with \textit{color inversion} (namely, SD-NR-GAN-III) achieves the best performance in all cases, while SD-NR-GAN-0 with any other transformation (i.e., \textit{rotation}, \textit{channel shuffle}, or \textit{all transformations}) does not necessarily improve the performance compared to SD-NR-GAN-0. Refer to Appendix~\ref{subsec:ablation_transformations}, where we discuss the effect of the transformations in SI-NR-GAN-II.

\begin{table}[tb]
  \centering
  \scriptsize{
    \begin{tabularx}{\columnwidth}{cYYYYYYYY} \toprule
      \multirow{2}{*}{\textbf{Model}} & (I) & (J) & (K) & (L) & (M) & (N) & (O) & (P)
      \\
      & \textbf{MGF} & \textbf{MGV} & \textbf{{\tiny A}+I} & \textbf{A+I} & \textbf{PF} & \textbf{PV} & \textbf{{\tiny A}+M} & \textbf{A+M}
      \\ \midrule
      SD-NR-GAN-II
      & 24.4 & 24.2 & 23.3 & 28.5 & 34.0 & 33.9 & 34.0 & 35.4
      \\
      \:\:- Implicit relational learning\:\:
      & 72.1 & 63.6 & 67.3 & 46.9 & 94.6 & 97.9 & 94.9 & 86.0
      \\ \midrule
      SD-NR-GAN-III
      & 37.5 & 33.4 & 33.5 & 33.9 & 53.1 & 55.1 & 52.4 & 47.2
      \\
      - Implicit relational learning
      & 75.7 & 67.5 & 69.0 & 47.3 & 100.6 & 102.0 & 98.3 & 80.5
      \\ \midrule
      GAN
      & 82.7 & 77.4 & 93.2 & 155.8 & 152.4 & 160.1 & 149.1  & 175.8
      \\ \bottomrule
    \end{tabularx}
  }
  \caption{
    \textbf{Comparison of FID on \textsc{CIFAR-10} with signal-dependent noises with and without implicit relational learning.} A smaller value is better. SD-NR-GAN-II without implicit relational learning (the fourth row) is equal to SI-NR-GAN-I. SD-NR-GAN-III without implicit relational learning (the sixth row) is equal to a variant of SI-NR-GAN-II, in which only \textit{color inversion} is used as a transformation. Herein, we report the median score for three random seeds.
  }
  \label{tab:ablation_implicit_learning}
  \vspace{-2mm}
\end{table}

\subsection{Ablation study on implicit relational learning in SD-NR-GAN-II and -III}
\label{subsec:ablation_implicit_learning}

We conducted an ablation study to clarify the effectiveness of implicit relational learning in SD-NR-GAN-II and -III for signal-dependent noises. Table~\ref{tab:ablation_implicit_learning} lists the results. SD-NR-GAN-II without implicit relational learning is equal to SI-NR-GAN-I, and SD-NR-GAN-III without implicit relational learning is equal to a variant of SI-NR-GAN-II, in which only \textit{color inversion} is used as a transformation. We found that, although SD-NR-GAN-II and -III without implicit relational learning (the fourth and sixth rows) outperform standard GAN in all cases, they are defeated by SD-NR-GAN-II and -III with implicit relational learning (the third and fifth rows) by a large margin in all cases (with a difference of over 10). These results verify the effectiveness of implicit relational learning for signal-dependent noises. We discuss the generality of SD-NR-GAN-II and -III for signal-independent noises in Appendix~\ref{subsec:generality_sd_nr_gans}.

\begin{table}[tb]
  \centering
  \scriptsize{
    \begin{tabularx}{\columnwidth}{cYYYYYYYY} \toprule
      \multirow{2}{*}{\textbf{Model}} & (A) & (B) & (C) & (D) & (E) & (F) & (G) & (H)
      \\
      & \textbf{AGF} & \textbf{AGV} & \textbf{LGF} & \textbf{LGV} & \textbf{U} & \textbf{MIX} & \textbf{BG} & \textbf{A+G}
      \\ \midrule
      SD-NR-GAN-II
      & 26.7 & 28.0 & 23.7 & 22.3 & 43.7 & 24.7 & 164.8 & 249.2
      \\
      \:\:- Implicit relational learning\:\:
      & 26.7 & 27.5 & 22.1 & 22.4 & 40.1 & 24.8 & 163.4 & 253.2
      \\ \midrule
      SD-NR-GAN-III
      & 30.9 & 31.0 & 24.1 & 23.9 & 32.7 & 27.9 & 35.8 & 46.5
      \\
      - Implicit relational learning
      & 29.5 & 30.3 & 22.3 & 22.0 & 32.1 & 26.8 & 33.6 & 44.3
      \\ \midrule
      GAN
      & 145.8 & 136.0 & 38.8 & 38.8 & 146.4 & 125.6 & 165.3 & 265.9
      \\ \bottomrule
    \end{tabularx}
  }
  \caption{
    \textbf{Comparison of FID on \textsc{CIFAR-10} with signal-independent noises with and without implicit relational learning.} A smaller value is better. The explanation of the models is provided in Table~\ref{tab:ablation_implicit_learning}. The median score for three random seeds is provided.
  }
  \label{tab:fid_sdnrgan_general}
  \vspace{-2mm}
\end{table}

\subsection{Generality analysis of SD-NR-GAN-II and -III}
\label{subsec:generality_sd_nr_gans}

In the main text, we demonstrate the effectiveness of SD-NR-GAN-II and -III for signal-dependent noises. However, as discussed in Sections~\ref{subsec:SD-NR-GAN-II} and \ref{subsec:SD-NR-GAN-III}, SD-NR-GAN-II and -III can also be applied to signal-independent noises because they can \textit{implicitly} learn the signal-noise relationship, i.e., learn $R({\bm x}, {\bm \sigma}) = {\bm x}$ in this case. To verify this claim, we examined the performance of SD-NR-GAN-II and -III on signal-independent noises (Figure~\ref{fig:noise_examples}(A)--(H)). To focus on an analysis of implicit relational learning, we compared SD-NR-GAN-II with SI-NR-GAN-I and SD-NR-GAN-III with a variant of SI-NR-GAN-II, in which only \textit{color inversion} is used as a transformation. Using these settings, the differences among these two models are whether $G_{\bm n}$ incorporates ${\bm z}_{\bm x}$ into an input, i.e., whether an implicit relational learning is conducted.

Table~\ref{tab:fid_sdnrgan_general} summarizes the results. We found that, although in both cases SD-NR-GANs (the third and fifth rows) are defeated by SI-NR-GANs (the fourth and sixth rows), the difference is relatively small (the worst degradation is 3.6 in SD-NR-GAN-II and 2.2 in SD-NR-GAN-III). Furthermore, we observed that SD-NR-GANs outperform standard GAN by a large margin. From these results, we conclude that, although the use of SI-NR-GANs is the best when it is known that the noise is signal-independent, using SD-NR-GAN-II and -III is also reasonable when the noise type is unknown. We discuss the limitations of SI-NR-GANs for signal-dependent noises in Appendix~\ref{subsec:ablation_implicit_learning}.

\begin{table}[tbh!]
  \centering
  \scriptsize{
    \begin{tabularx}{\columnwidth}{cYYYY} \toprule
      \multirow{3}{*}{\textbf{Method}} & \multicolumn{3}{c}{\textbf{\textsc{LSUN Bedroom}}} & \textbf{\textsc{FFHQ}}
      \\ \cmidrule(lr){2-4} \cmidrule{5-5}
      & (A) & (B) & (G) & (A)
      \\
      & \textbf{AGF} & \textbf{AGV} & \textbf{BG} & \textbf{AGF}
      \\ \midrule
      \multirow{2}{*}{GN2GC w/ SI-NR-GAN-I}
      & \textbf{32.36} & \textbf{32.47} & 20.74 & \textbf{31.34}
      \\
      & (\textbf{13.8}) & (\textbf{14.2}) & (128.2) & (\textbf{35.7})
      \\ \midrule
      \multirow{2}{*}{GN2GC w/ SI-NR-GAN-II}
      & 31.49 & 31.70 & \textbf{26.61} & 30.81
      \\
      & (15.7) & (16.8) & (\textbf{10.8}) & (37.1)
      \\ \midrule
      Noisy
      & 20.52 & 21.01 & 20.72 & 20.60 
      \\ \bottomrule
      \multicolumn{5}{c}{\multirow{2}{*}{\footnotesize{(a) Signal-independent noise}}}
      \\
      \multicolumn{5}{c}{}
      \\ \toprule
      \multirow{3}{*}{\textbf{Method}} & \multicolumn{3}{c}{\textbf{\textsc{LSUN Bedroom}}} & \textbf{\textsc{FFHQ}}
      \\ \cmidrule(lr){2-4} \cmidrule{5-5}
      & (I) & (L) & (M) & (I)
      \\
      & \textbf{MGF} & \textbf{A+I} & \textbf{PF} & \textbf{MGF}
      \\ \midrule
      \multirow{2}{*}{GN2GC w/ SD-NR-GAN-I}
      & \textbf{36.01} & 25.75 & \textbf{31.08} & \textbf{35.69}
      \\
      & (\textbf{11.6}) & (55.4) & (\textbf{23.3}) & (\textbf{26.5})
      \\ \midrule
      \multirow{2}{*}{GN2GC w/ SD-NR-GAN-II}
      & 35.43 & \textbf{31.62} & 30.00 & 34.58
      \\
      & (21.7) & (\textbf{15.0}) & (42.8) & (49.0)
      \\ \midrule
      \multirow{2}{*}{\:\:GN2GC w/ SD-NR-GAN-III\:\:}
      & 30.16 & 28.47 & 22.92 & 29.91
      \\
      & (50.7) & (53.1) & (138.6) & (37.2)
      \\ \midrule
      Noisy
      & 25.49 & 19.37 & 18.30 & 26.58
      \\ \bottomrule
      \multicolumn{5}{c}{\multirow{2}{*}{\footnotesize{(b) Signal-dependent noise}}}
      \\
    \end{tabularx}
  }
  \vspace{1mm}
  \caption{
    \textbf{Comparison of PSNR on \textsc{LSUN Bedroom} and \textsc{FFHQ} using different GN2GCs.} A larger value is better. In the parenthesis, we show the FID of NR-GANs provided in Table~\ref{tab:fid_complex} (a smaller value is better). Bold font indicates the best score.
  }
  \label{tab:psnr_gn2gc}
\end{table}

\subsection{Detailed analysis on GN2GC}
\label{subsec:analysis_gn2gc}

In the main text, we employed those generators achieving the best FID for GN2GC, as described in Section~\ref{subsec:application}. In this section, we provide other case results.

Table \ref{tab:psnr_gn2gc}(a) shows the results for signal-independent noises. We can see that most of the models improve the PSNR compared to the original noisy input. An exception is GN2GC with SI-NR-GAN-I in \textsc{LSUN Bedroom} under Brown Gaussian noise (G). In this case, SI-NR-GAN-I fails to learn a clean image generator because the noise is beyond the assumption of SI-NR-GAN-I. This causes a difficulty in learning a denoiser, and the PSNR is almost the same as the noisy input. When comparing GN2GC with SI-NR-GAN-I and GN2GC with SI-NR-GAN-II, we can see that GN2GC obtains a better PSNR when using SI-NR-GAN, which achieves a better FID.

\begin{figure}[tb!]
  \centering
  \includegraphics[width=\columnwidth]{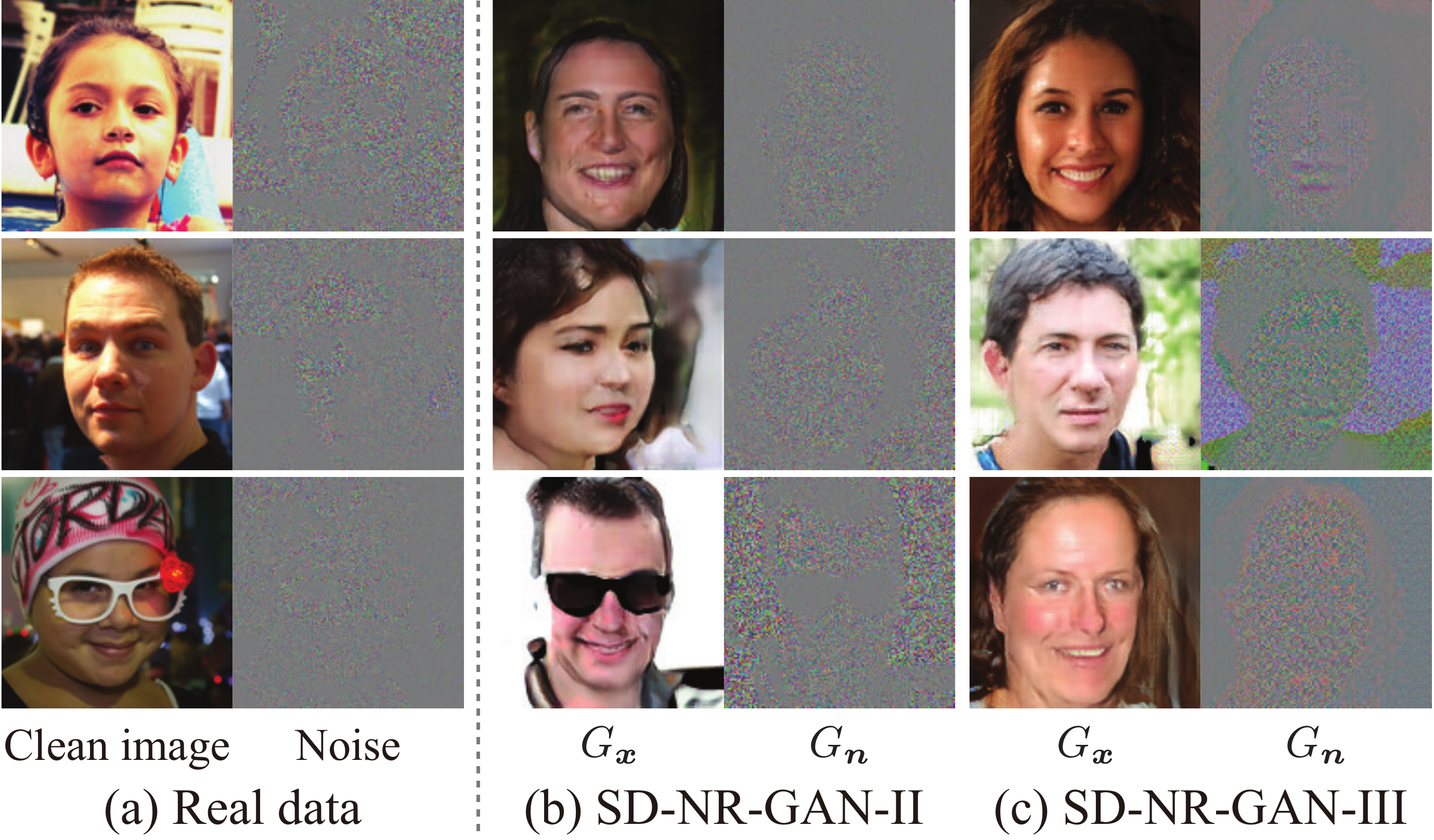}
  \caption{
    \textbf{Examples of images and noises generated for \textsc{FFHQ} with multiplicative Gaussian noise (Figure~\ref{fig:noise_examples}(I)).} The better $G_{\bm x}$ is obtained for SD-NR-GAN-III, while the better $G_{\bm n}$ is learned with SD-NR-GAN-II.
  }
  \label{fig:examples_ffhq_detail}
  \vspace{-3mm}
\end{figure}

Table~\ref{tab:psnr_gn2gc}(b) shows the results for signal-dependent noises. We can observe that all models improve the PSNR compared to the original noisy input even when using the worst FID model (GN2GC with SD-NR-GAN-III in \textsc{LSUN Bedroom} with Poisson noise (M)). Similar to the results for signal-independent noises, when comparing GN2GC with SD-NR-GAN-I, GN2GC with SD-NR-GAN-II, and GN2GC with SD-NR-GAN-III, we found that GN2GC tends to obtain a better PSNR when using SD-NR-GAN, which achieves the best FID. An exception to this is GN2GC with SD-NR-GAN-II in \textsc{FFHQ} under multiplicative Gaussian noise (I). In this case, SD-NR-GAN-II is worse than SD-NR-GAN-III in terms of the FID, whereas GN2GC with SD-NR-GAN-II outperforms GN2GC with SD-NR-GAN-III. This is because the better $G_{\bm x}$ is obtained for SD-NR-GAN-III, while the better $G_{\bm n}$ is learned with SD-NR-GAN-II, as shown in Figure~\ref{fig:examples_ffhq_detail}. We show examples of denoised images in Figures~\ref{fig:examples_complex_denoised_si} and \ref{fig:examples_complex_denoised_sd}.

\newpage
\section{Examples of generated and denoised images}
\label{sec:examples}

In Figures~\ref{fig:examples_cifar10_si}--\ref{fig:examples_complex_denoised_sd}, we show examples of images generated or denoised by the models described in Sections~\ref{subsec:comprehensive_study}--\ref{subsec:application}.

\subsection{Examples of generated images for Section~\ref{subsec:comprehensive_study}}
\label{subsec:examples_comprehensive_study}

\begin{itemize}
  \setlength{\parskip}{1pt}
  \setlength{\itemsep}{1pt}
\item Figure~\ref{fig:examples_cifar10_si}:
  Examples of images generated for \textsc{CIFAR-10} with signal-independent noises
\item Figure~\ref{fig:examples_cifar10_sd}:
  Examples of images generated for \textsc{CIFAR-10} with signal-dependent noises
\end{itemize}

\begin{figure*}[t]
  \centering
  \includegraphics[height=0.94\textheight]{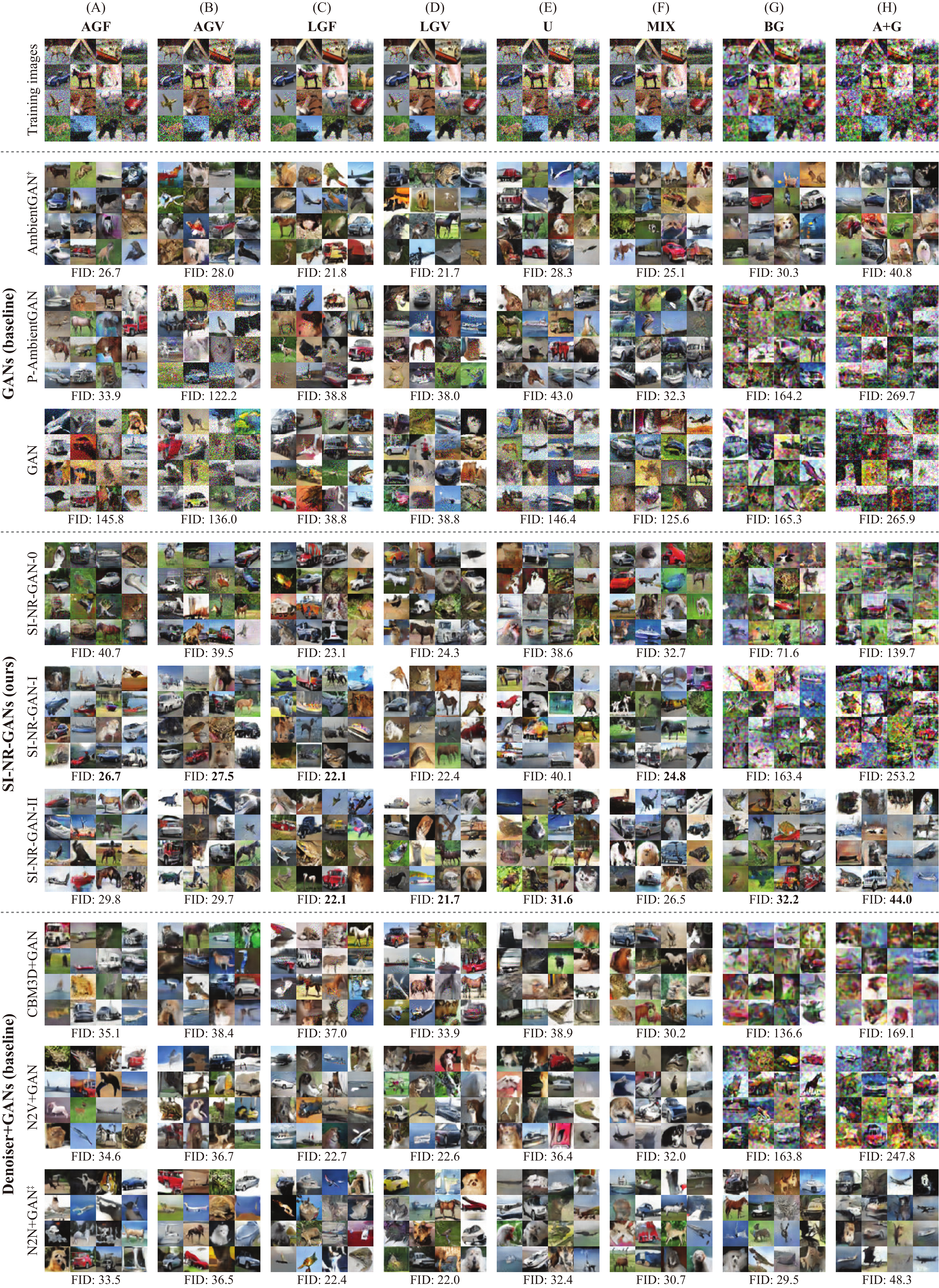}
  \caption{
    \textbf{Examples of images generated for \textsc{CIFAR-10} with signal-independent noises.} We show $4 \times 4$ samples for each case. AmbientGAN$^{\dag}$ is trained with the ground-truth noise models. N2N+GAN$^{\ddag}$ uses noisy image pairs when training N2N. The other models are trained using only noisy images (not including pairs) without complete noise information.
  }
  \label{fig:examples_cifar10_si}
\end{figure*}

\begin{figure*}[t]
  \centering
  \includegraphics[height=0.94\textheight]{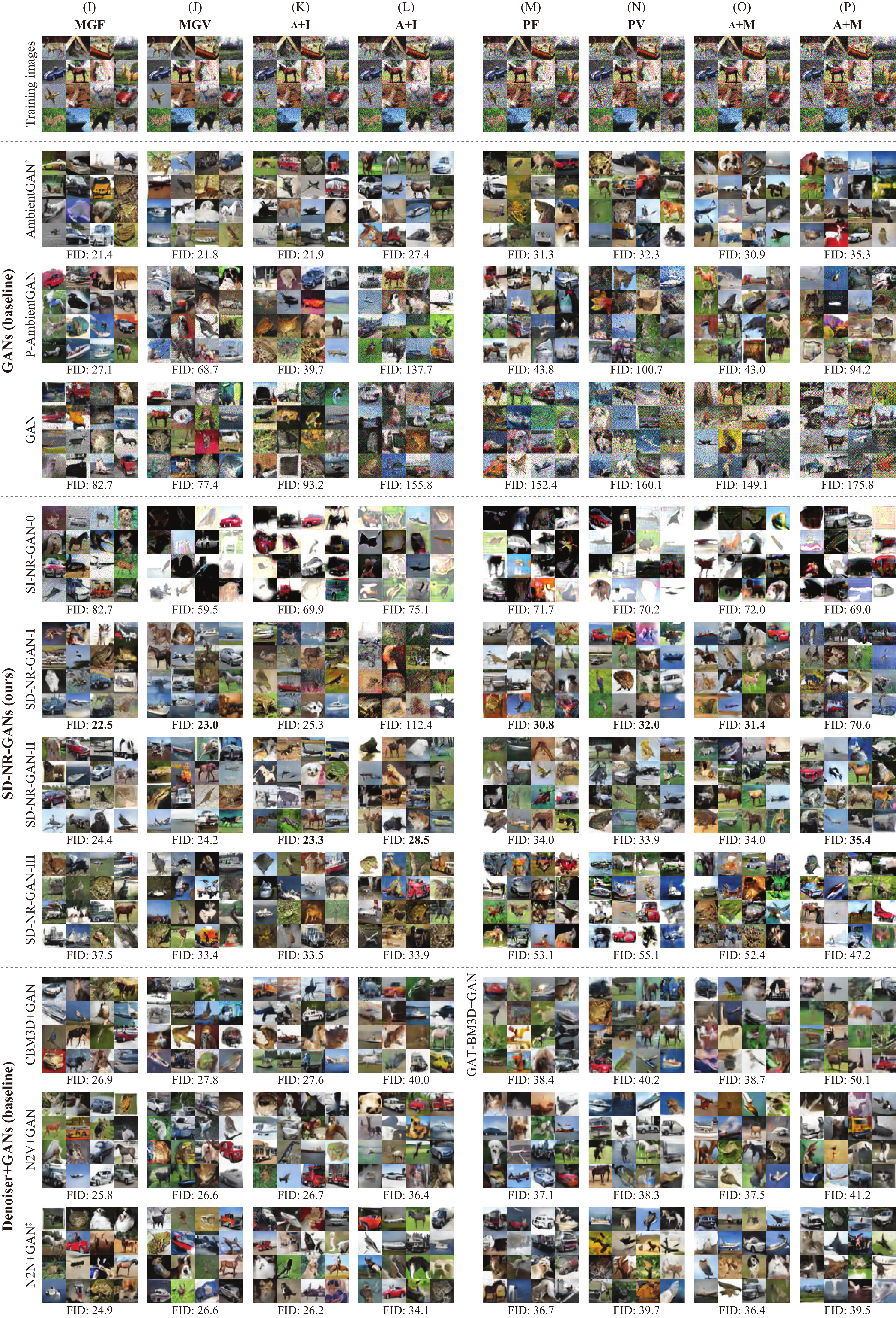}
  \caption{
    \textbf{Examples of images generated for \textsc{CIFAR-10} with signal-dependent noises.} Here, $4 \times 4$ samples are shown for each case. AmbientGAN$^{\dag}$ is trained with the ground-truth noise models. N2N+GAN$^{\ddag}$ uses noisy image pairs when training N2N. The other models are trained using only noisy images (not including pairs) without full knowledge of the noise.
  }
  \label{fig:examples_cifar10_sd}
\end{figure*}

\subsection{Examples of generated images for Section~\ref{subsec:complex}}
\label{subsec:examples_complex}

\begin{itemize}
  \setlength{\parskip}{1pt}
  \setlength{\itemsep}{1pt}
\item Figure~\ref{fig:examples_complex_si}:
  Examples of images generated for \textsc{LSUN Bedroom} and \textsc{FFHQ} with signal-independent noises
\item Figure~\ref{fig:examples_complex_si_with_noise}:
  Examples of images and noises generated for \textsc{LSUN Bedroom} and \textsc{FFHQ} with signal-independent noises
\item Figure~\ref{fig:examples_complex_sd}:
  Examples of images generated for \textsc{LSUN Bedroom} and \textsc{FFHQ} with signal-dependent noises
\item Figure~\ref{fig:examples_complex_sd_with_noise}:
  Examples of images and noises generated for \textsc{LSUN Bedroom} and \textsc{FFHQ} with signal-dependent noises
\end{itemize}

\begin{figure*}[t]
  \centering
  \includegraphics[height=0.9\textheight]{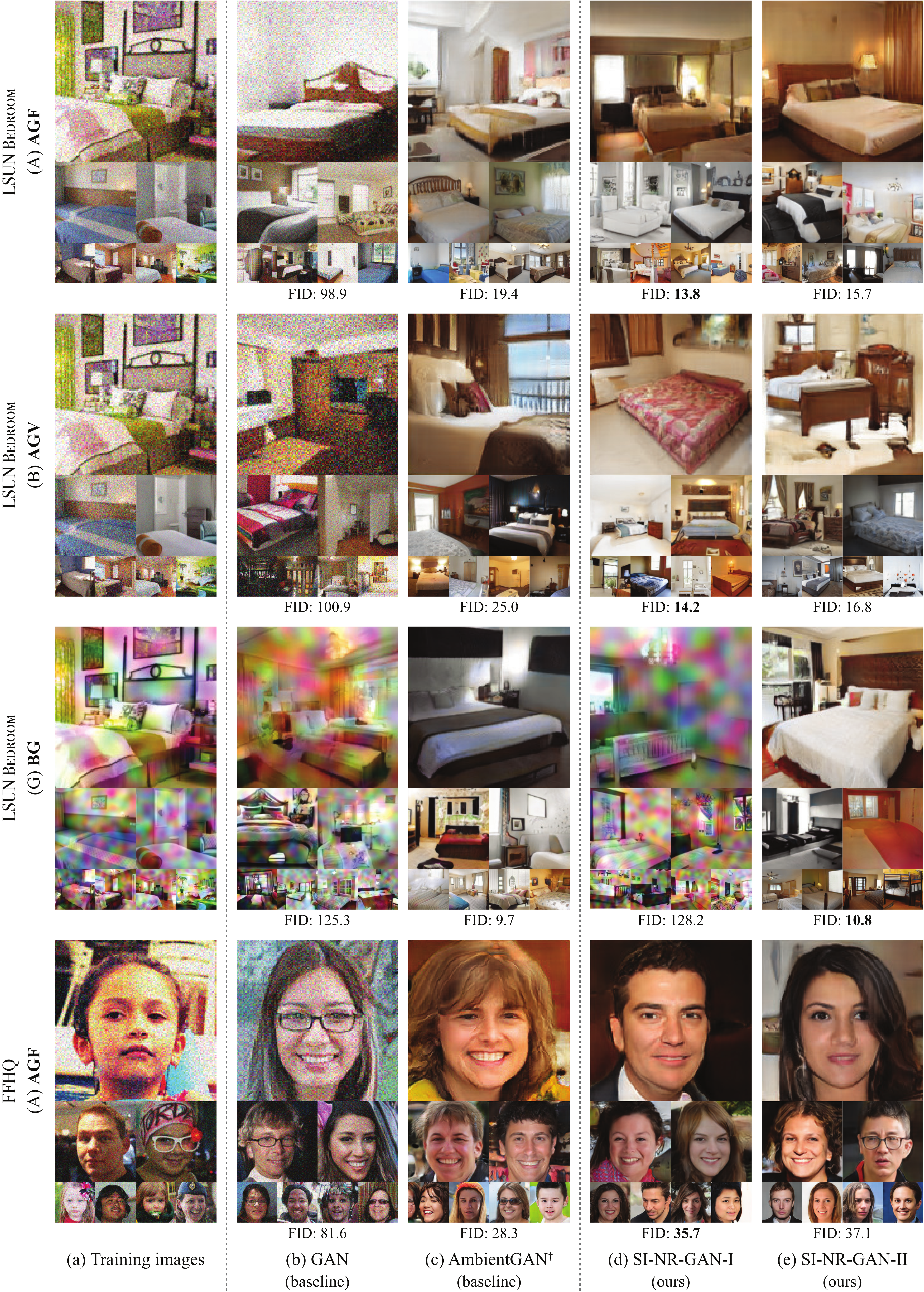}
  \caption{
    \textbf{Examples of images generated for \textsc{LSUN Bedroom} and \textsc{FFHQ} with signal-independent noises.} We show $1 + 2 + 4$ samples for each case. AmbientGAN$^{\dag}$ is trained with the ground-truth noise models. The other models are trained without complete noise information.
  }
  \label{fig:examples_complex_si}
\end{figure*}

\begin{figure*}[t]
  \centering
  \includegraphics[height=0.75\textheight]{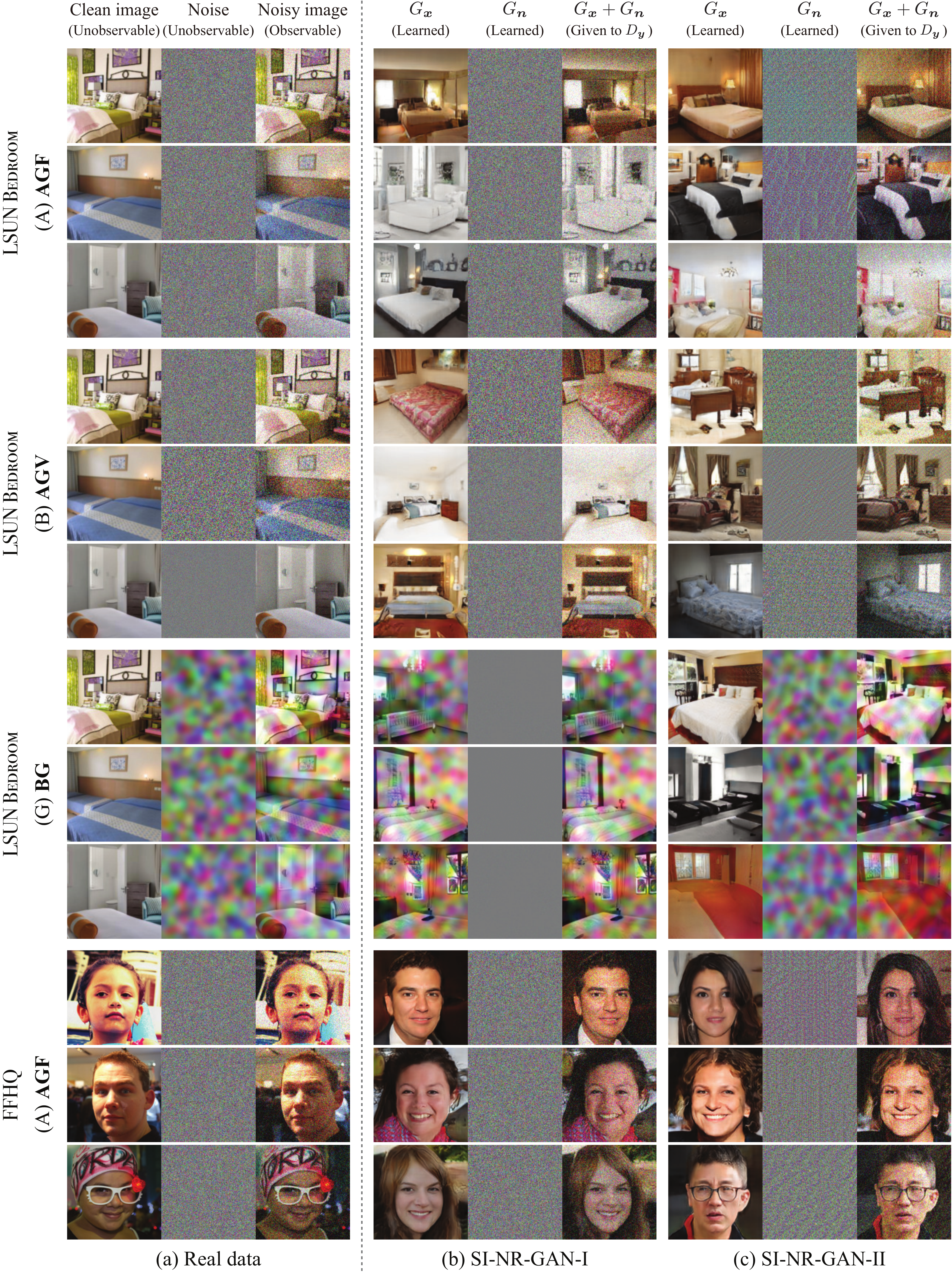}
  \caption{
    \textbf{Examples of images and noises generated for \textsc{LSUN Bedroom} and \textsc{FFHQ} with signal-independent noises.} Here, three samples are shown for each case. (a) During the training, only noisy images (third column) are given and clean images (first column) and ground-truth noises (second column) are not observable. (b)(c) Our proposed SI-NR-GANs learn to generate clean images (fourth and seventh columns) and noises (fifth and eighth columns) simultaneously. Their combinations (sixth and ninth columns) are given to $D_{\bm y}$ and are compared with real noisy images (third column).
  }
  \label{fig:examples_complex_si_with_noise}
\end{figure*}

\begin{figure*}[t]
  \centering
  \includegraphics[height=0.9\textheight]{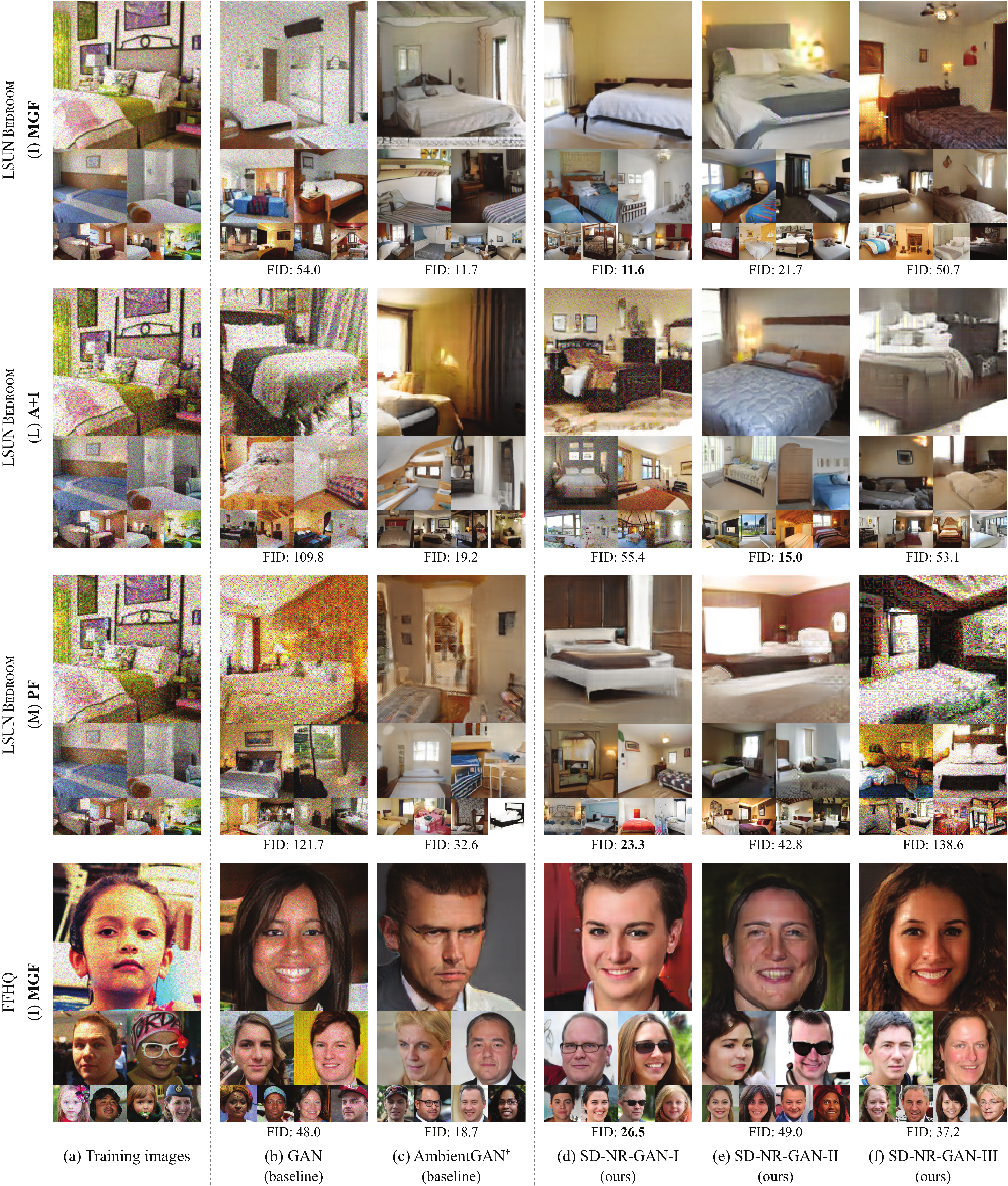}
  \caption{
    \textbf{Examples of images generated for \textsc{LSUN Bedroom} and \textsc{FFHQ} with signal-dependent noises.} We show $1 + 2 + 4$ samples for each case. AmbientGAN$^{\dag}$ is trained with the ground-truth noise models. The other models are trained without full knowledge of the noise.
  }
  \label{fig:examples_complex_sd}
\end{figure*}

\begin{figure*}[t]
  \centering
  \includegraphics[height=0.75\textheight]{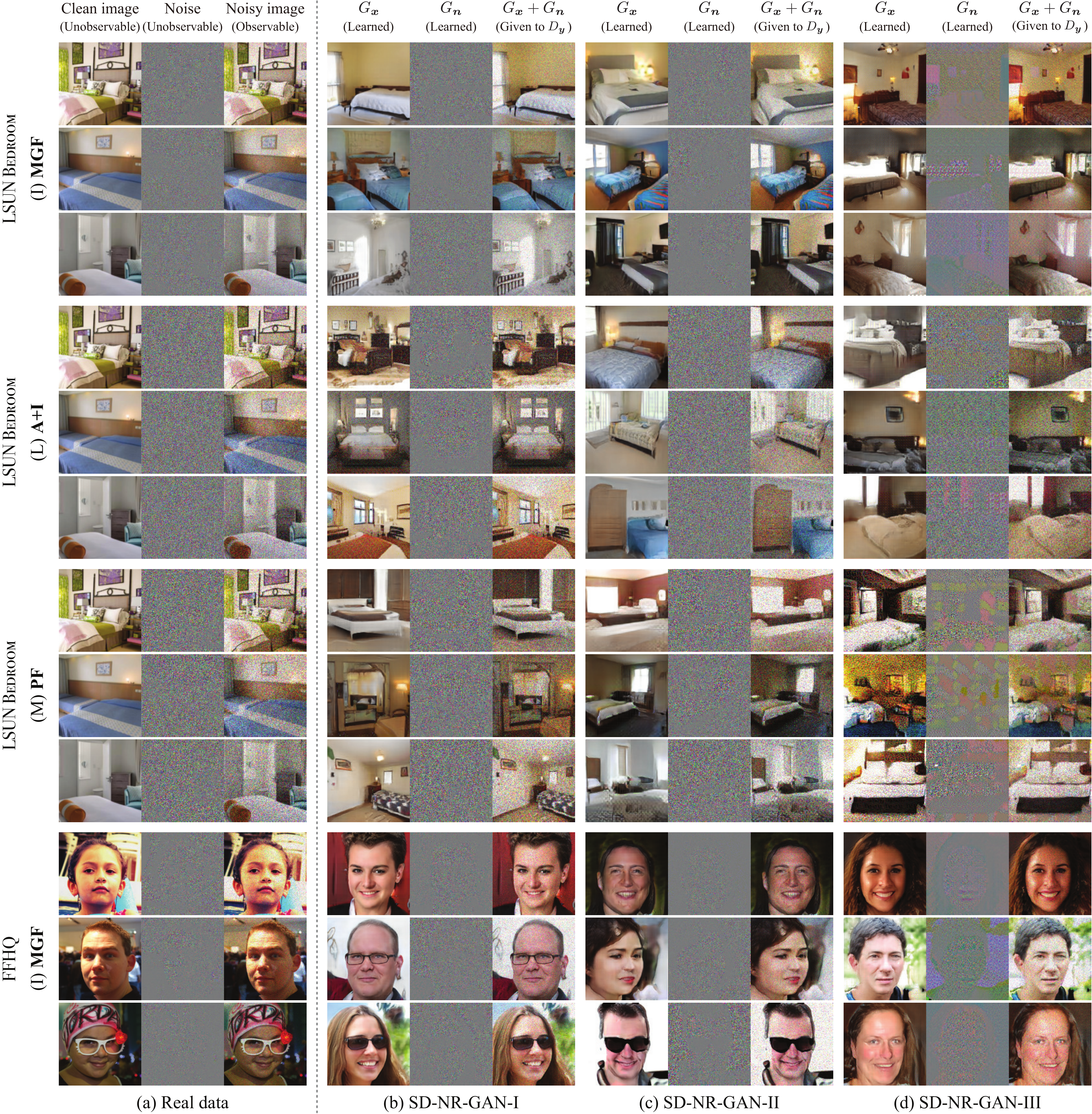}
  \caption{
    \textbf{Examples of images and noises generated for \textsc{LSUN Bedroom} and \textsc{FFHQ} with signal-dependent noises.} We show three samples for each case. (a) During the training, only noisy images (third column) are given and clean images (first column) and ground-truth noises (second column) are not observable. (b)(c)(d) Our proposed SD-NR-GANs learn to generate clean images (fourth, seventh, and tenth columns) and noises (fifth, eighth, and eleventh columns) at the same time. Their combinations (sixth, ninth, and twelfth columns) are provided to $D_{\bm y}$ and are compared with real noisy images (third column).
  }
  \label{fig:examples_complex_sd_with_noise}
\end{figure*}

\subsection{Examples of denoised images for Section~\ref{subsec:application}}
\label{subsec:examples_application}

\begin{itemize}
  \setlength{\parskip}{1pt}
  \setlength{\itemsep}{1pt}
\item Figure~\ref{fig:examples_complex_denoised_si}:
  Examples of denoised images for \textsc{LSUN Bedroom} and \textsc{FFHQ} under signal-independent noises
\item Figure~\ref{fig:examples_complex_denoised_sd}:
  Examples of denoised images for \textsc{LSUN Bedroom} and \textsc{FFHQ} under signal-dependent noises
\end{itemize}

\begin{figure*}[t]
  \centering
  \includegraphics[width=0.9\textwidth]{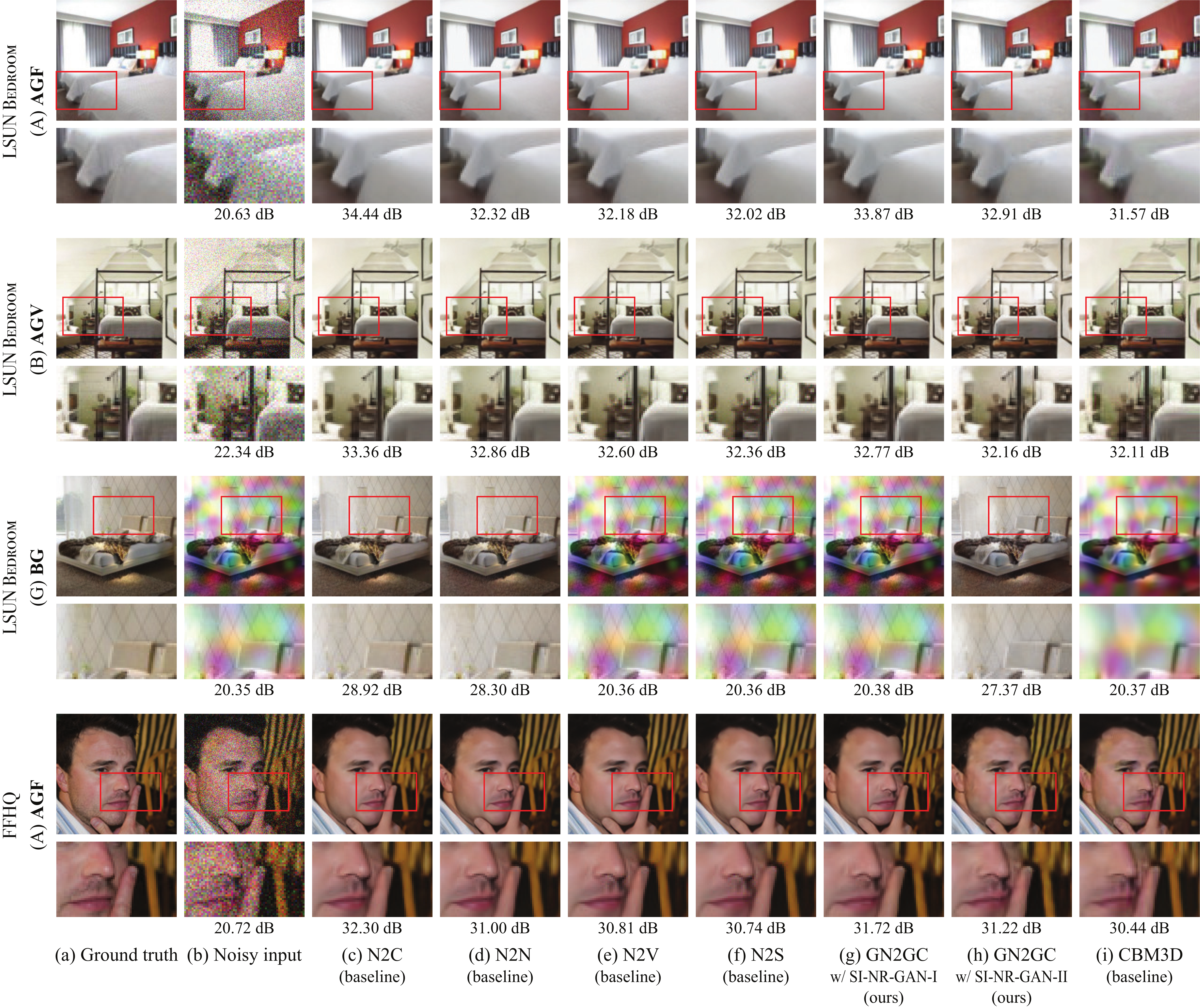}
  \caption{
    \textbf{Examples of denoised images for \textsc{LSUN Bedroom} and \textsc{FFHQ} under signal-independent noises.} Numbers below images indicate the PSNR for each image.
  }
  \label{fig:examples_complex_denoised_si}
\end{figure*}

\begin{figure*}[t]
  \centering
  \includegraphics[width=1.0\textwidth]{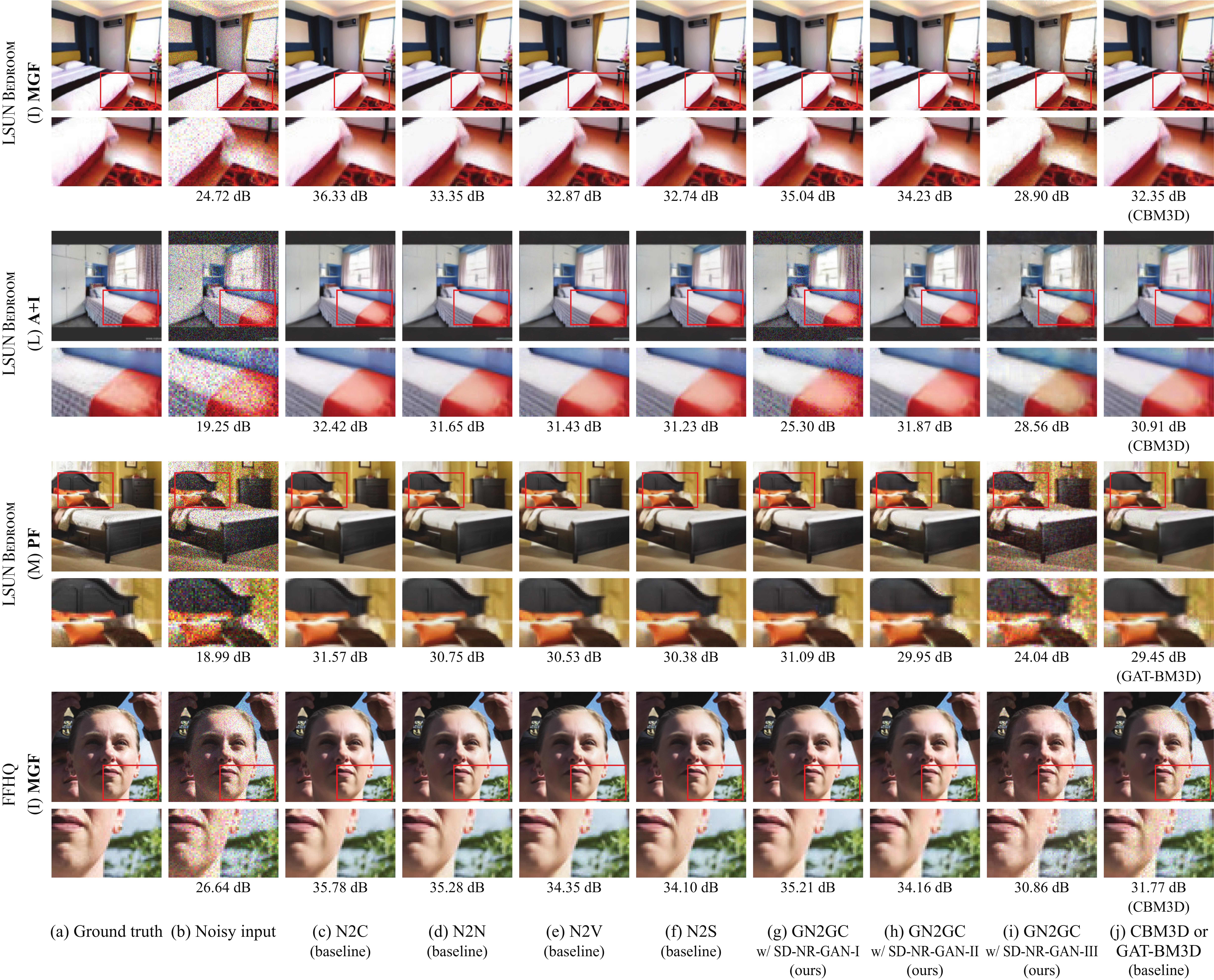}
  \caption{
    \textbf{Examples of denoised images for \textsc{LSUN Bedroom} and \textsc{FFHQ} under signal-dependent noises.} Numbers below images indicate the PSNR for each image.
  }
  \label{fig:examples_complex_denoised_sd}
\end{figure*}

\clearpage
\section{Implementation details}
\label{sec:details}

\smallskip\noindent\textbf{Notation.}
In the description of the network architectures, we apply the following notations.
\begin{itemize}
  \setlength{\parskip}{1pt}
  \setlength{\itemsep}{1pt}
\item Linear:
  Linear layer
\item Conv:
  Convolutional layer
\item ReLU:
  Rectified unit~\cite{VNairICML2010}
\item LReLU:
  Leaky rectified unit~\cite{AMaasICML2013,BXuICMLW2015}
\item ResBlock:
  Residual block~\cite{KHeCVPR2016}
\end{itemize}

In the description of the training settings, we use the following notations. We used an Adam optimizer~\cite{DPKingmaICLR2015} during all experiments.
\begin{itemize}
  \setlength{\parskip}{1pt}
  \setlength{\itemsep}{1pt}
\item $\alpha$:
  Learning rate of Adam
\item $\beta_{1}$:
  The first-order momentum parameter of Adam
\item $\beta_{2}$:
  The second-order momentum parameter of Adam
\end{itemize}

\subsection{Details on Section~\ref{subsec:comprehensive_study}}
\label{subsec:detail_comprehensive_study}

\subsubsection{Noise details}
\label{subsubsec:detail_comprehensive_study_noise}

As described in Section~\ref{subsec:comprehensive_study}, we tested 16 noises, shown in Figure~\ref{fig:noise_examples}(A)--(P), the details of which are provided in the caption.

\subsubsection{Details of GANs}
\label{subsubsec:detail_comprehensive_study_gan}

In this section, we provide the details of GANs (AmbientGAN, P-AmbientGAN, GAN, SI-NR-GANs, and SD-NR-GANs) described in Section~\ref{subsec:comprehensive_study}.

\smallskip\noindent\textbf{Network architectures.}
Table~\ref{tab:net_resnet32} shows the generator and discriminator network architectures. We used the ResNet architectures~\cite{KHeCVPR2016}. Following the study on a real gradient penalty regularization ($R_1$ regularization)~\cite{LMeschederICML2018}, which we used as a GAN regularization, we multiplied the output of the ResNet blocks with 0.1 and did not use batch normalization~\cite{SIoffeICML2015}. In NR-GANs, we used similar architectures for $G_{\bm x}$ and $G_{\bm n}$. The number of dimensions of the latent vector $d_{\bm z}$ in the generator was set to 128, except for $G_{\bm n}$ in SD-NR-GAN-II and -III, where $d_{\bm z}$ was set to 256 because the noise generator $G_{\bm n}({\bm z}_{\bm n}, {\bm z}_{\bm x})$ in these models receives an image latent vector ${\bm z}_{\bm x}$ along with a noise latent vector ${\bm z}_{\bm n}$. The images were normalized within the range of $[-1, 1]$.

\smallskip\noindent\textbf{Training settings.}
As a GAN objective function, we used a non-saturating GAN loss~\cite{IGoodfellowNIPS2014} with $R_1$ regularization~\cite{LMeschederICML2018}. We set the trade-off parameter for $R_1$ regularization to $10$. With NR-GANs, we additionally applied a diversity-sensitive regularization (DS regularization)~\cite{DYangICLR2019} with a trade-off parameter of 0.02. We discuss the effect of this parameter in Appendix~\ref{subsec:effect_ds_regularization}. We trained the networks for $200k$ iterations using the Adam optimizer~\cite{DPKingmaICLR2015} with $\alpha = 0.0002$, $\beta_1 = 0$, $\beta_2 = 0.99$, and a batch size of 64. We updated the generator and discriminator alternatively. Similar to prior studies~\cite{TKarrasICLR2018,LMeschederICML2018}, we used an exponential moving average with a decay of 0.999 over the weights to produce the final generator.

\begin{table}[tb]
  \centering
  \footnotesize{
    \begin{tabularx}{0.7\columnwidth}{Y} \toprule
      \textbf{Generator} $G({\bm z})$
      \\ \midrule
      ${\bm z} \in \mathbb{R}^{d_{\bm z}}$
      \\ \midrule
      Linear $\rightarrow$ $4 \times 4 \times 128$
      \\ \midrule
      ResBlock up $128$
      \\ \midrule
      ResBlock up $128$
      \\ \midrule
      ResBlock up $128$
      \\ \midrule
      ReLU, $3 \times 3$ Conv $3$
      \\ \midrule
      Tanh
      \\ \bottomrule
      \multicolumn{1}{c}{}
      \\ \toprule
      \textbf{Discriminator} $D({\bm y})$
      \\ \midrule
      ${\bm y} \in \mathbb{R}^{32 \times 32 \times 3}$
      \\ \midrule
      ResBlock down 128
      \\ \midrule
      ResBlock down 128
      \\ \midrule
      ResBlock 128
      \\ \midrule
      ResBlock 128
      \\ \midrule
      ReLU, Global average pooling
      \\ \midrule
      Linear $\rightarrow$ 1
      \\ \bottomrule
    \end{tabularx}
  }
  \vspace{1mm}
  \caption{
    \textbf{Generator and discriminator architectures for \textsc{CIFAR-10}.}
  }
  \label{tab:net_resnet32}
\end{table}

\subsubsection{Details of denoisers}
\label{subsubsec:detail_comprehensive_study_denoiser}

In this section, we describe the details of the denoisers (N2N and N2V) used in Section~\ref{subsec:comprehensive_study}.

\smallskip\noindent\textbf{Network architectures.}
Table~\ref{tab:net_denoiser} shows the denoiser network architecture. We used the U-net architecture~\cite{ORonnebergerMICCAI2015}. This is the same as that used in the study on N2N~\cite{JLehtinenICML2018}. The input images were normalized within the range of $[-0.5, 0.5]$.

\smallskip\noindent\textbf{Training settings.}
We trained the network for $200k$ iterations using the Adam optimizer~\cite{DPKingmaICLR2015} with $\alpha = 0.0003$, $\beta_1 = 0.9$, and $\beta_2 = 0.99$. The learning rate was kept constant during the training except for the last $30\%$ iterations, where the learning rate was smoothly ramped down to zero. Following the study on N2N~\cite{JLehtinenICML2018}, we used a batch size of 4 for N2N. Just like the observation in the study of N2V~\cite{AKrullCVPR2019}, we found that N2V works better with a larger batch size. Hence, we used a batch size of 64 for N2V. Following the study on N2V~\cite{AKrullCVPR2019}, we manipulated 64 pixels per input image and used a uniform pixel selection (UPS) with a kernel size of $5 \times 5$ as a masking method.

\begin{table}[tb]
  \centering
  \footnotesize{
    \begin{tabularx}{0.85\columnwidth}{cY} \toprule
      \multicolumn{2}{c}{\textbf{Denoiser}}
      \\ \midrule
      \textbf{Name} & \textbf{Layer}
      \\ \midrule
      \textsc{input} & ${\bm y} \in \mathbb{R}^{H \times W \times 3}$
      \\ \midrule
      \textsc{enc{\_}conv0} & $3 \times 3$ Conv $48$, LReLU
      \\
      \textsc{enc{\_}conv1} & $3 \times 3$ Conv $48$, LReLU
      \\
      \textsc{pool1} & $2 \times 2$ Maxpool
      \\ \midrule
      \textsc{enc{\_}conv2} & $3 \times 3$ Conv $48$, LReLU
      \\
      \textsc{pool2} & $2 \times 2$ Maxpool
      \\ \midrule
      \textsc{enc{\_}conv3} & $3 \times 3$ Conv $48$, LReLU
      \\
      \textsc{pool3} & $2 \times 2$ Maxpool
      \\ \midrule
      \textsc{enc{\_}conv4} & $3 \times 3$ Conv $48$, LReLU
      \\
      \textsc{pool4} & $2 \times 2$ Maxpool
      \\ \midrule
      \textsc{enc{\_}conv5} & $3 \times 3$ Conv $48$, LReLU
      \\
      \textsc{pool5} & $2 \times 2$ Maxpool
      \\ \midrule
      \textsc{enc{\_}conv6} & $3 \times 3$ Conv $48$, LReLU
      \\ \midrule
      \textsc{Upsample5} & $2 \times 2$ Upsample
      \\
      \textsc{concat5} & Concatenate output of \textsc{pool4}
      \\
      \textsc{dec{\_}conv5a} & $3 \times 3$ Conv $96$, LReLU
      \\
      \textsc{dec{\_}conv5b} & $3 \times 3$ Conv $96$, LReLU
      \\ \midrule
      \textsc{Upsample4} & $2 \times 2$ Upsample
      \\
      \textsc{concat4} & Concatenate output of \textsc{pool3}
      \\
      \textsc{dec{\_}conv4a} & $3 \times 3$ Conv $96$, LReLU
      \\
      \textsc{dec{\_}conv4b} & $3 \times 3$ Conv $96$, LReLU
      \\ \midrule
      \textsc{Upsample3} & $2 \times 2$ Upsample
      \\
      \textsc{concat3} & Concatenate output of \textsc{pool2}
      \\
      \textsc{dec{\_}conv3a} & $3 \times 3$ Conv $96$, LReLU
      \\
      \textsc{dec{\_}conv3b} & $3 \times 3$ Conv $96$, LReLU
      \\ \midrule
      \textsc{Upsample2} & $2 \times 2$ Upsample
      \\
      \textsc{concat2} & Concatenate output of \textsc{pool1}
      \\
      \textsc{dec{\_}conv2a} & $3 \times 3$ Conv $96$, LReLU
      \\
      \textsc{dec{\_}conv2b} & $3 \times 3$ Conv $96$, LReLU
      \\ \midrule
      \textsc{Upsample1} & $2 \times 2$ Upsample
      \\
      \textsc{concat1} & Concatenate output of \textsc{input}
      \\
      \textsc{dec{\_}conv1a} & $3 \times 3$ Conv $64$, LReLU
      \\
      \textsc{dec{\_}conv1b} & $3 \times 3$ Conv $32$, LReLU
      \\
      \:\:\textsc{dec{\_}conv1c}\:\: & $3 \times 3$ Conv $3$
      \\ \bottomrule
    \end{tabularx}
  }
  \vspace{1mm}
  \caption{
    \textbf{Denoiser architecture.} This network is fully convolutional; therefore, it can take an arbitrary-size image as an input.
  }
  \label{tab:net_denoiser}
\end{table}

\subsubsection{Evaluation details}
\label{subsubsec:detail_comprehensive_study_evaluation}

As discussed in the main text, we used the Fr\'{e}chet inception distance (FID)~\cite{MHeuselNIPS2017} as an evaluation metric because its validity has been demonstrated in recent large-scale studies on GANs~\cite{MLucicNeurIPS2018,KKurachICML2019}, and because its sensitivity to the noise has also been shown \cite{MHeuselNIPS2017}. This metric measures the 2-Wasserstein distance between a real distribution $p^r$ and a generative distribution $p^g$ using the following equation:
\begin{flalign}
  \label{eqn:fid}
  d^2(p^r, p^g) = & \: \| {\bm m}^r - {\bm m}^g \|^2_2
  \nonumber \\
  + & \: \text{Tr}({\bm C}^r + {\bm C}^g - 2 ({\bm C}^r {\bm C}^g)^{\frac{1}{2}}),
\end{flalign}
where $\{ {\bm m}^r, {\bm C}^r \}$ and $\{ {\bm m}^g, {\bm C}^g \}$ denote the mean and covariance of the final feature vectors of the inception model calculated over real and generated samples, respectively. When calculating the FID, we used $10k$ real test samples and $10k$ generated samples, following the suggestion
from previous large-scale studies on GANs~\cite{MLucicNeurIPS2018,KKurachICML2019}.

\begin{table}[tb]
  \centering
  \footnotesize{
    \begin{tabularx}{0.7\columnwidth}{Y} \toprule
      \textbf{Generator} $G({\bm z})$
      \\ \midrule
      ${\bm z} \in \mathbb{R}^{d_{\bm z}}$
      \\ \midrule
      Linear $\rightarrow$ $4 \times 4 \times 1024$
      \\ \midrule
      ResBlock up $1024$
      \\ \midrule
      ResBlock up $512$
      \\ \midrule
      ResBlock up $256$
      \\ \midrule
      ResBlock up $128$
      \\ \midrule
      ResBlock up $64$
      \\ \midrule
      ReLU, $3 \times 3$ Conv $3$
      \\ \midrule
      Tanh
      \\ \bottomrule
      \multicolumn{1}{c}{}
      \\ \toprule
      \textbf{Discriminator} $D({\bm y})$
      \\ \midrule
      ${\bm y} \in \mathbb{R}^{128 \times 128 \times 3}$
      \\ \midrule
      ResBlock down 64
      \\ \midrule
      ResBlock down 128
      \\ \midrule
      ResBlock down 256
      \\ \midrule
      ResBlock down 512
      \\ \midrule
      ResBlock down 1024
      \\ \midrule
      ResBlock 1024
      \\ \midrule
      ReLU, Global average pooling
      \\ \midrule
      Linear $\rightarrow$ 1
      \\ \bottomrule
    \end{tabularx}
  }
  \vspace{1mm}
  \caption{
    \textbf{Generator and discriminator architectures for \textsc{LSUN Bedroom} and \textsc{FFHQ}.}
  }
  \label{tab:net_resnet128}
  \vspace{-2mm}
\end{table}

\subsection{Details on Section~\ref{subsec:complex}}
\label{subsec:detail_complex}

\subsubsection{Noise details}
\label{subsubsec:detail_complex_noise}

As described in Section~\ref{subsec:complex}, we selected six noises for \textsc{LSUN Bedroom} and two noises for \textsc{FFHQ}, such that they include variations. The noise parameters are the same as those described in Section~\ref{subsec:comprehensive_study} except for Brown Gaussian noise (G) where a $31 \times 31$ Gaussian filter is used instead of a $5 \times 5$ Gaussian filter.

\subsubsection{Details of GANs}
\label{subsubsec:detail_complex_gan}

In this section, we provide the details regarding GANs (AmbientGAN, GAN, SI-NR-GANs, and SD-NR-GANs) described in Section~\ref{subsec:complex}.

\smallskip\noindent\textbf{Network architectures.}
Table~\ref{tab:net_resnet128} shows the generator and discriminator network architectures. We used the ResNet architectures~\cite{KHeCVPR2016}. Following a previous study on $R_1$ regularization~\cite{LMeschederICML2018}, which we used as a GAN regularization, we multiplied the output of the ResNet blocks with 0.1 and did not use batch normalization~\cite{SIoffeICML2015}. In NR-GANs, we used similar architectures for $G_{\bm x}$ and $G_{\bm n}$. The number of dimensions of the latent vector $d_{\bm z}$ in the generator was set to 256, except for $G_{\bm n}$ in SD-NR-GAN-II and -III, where $d_{\bm z}$ was set to 512 because the noise generator $G_{\bm n}({\bm z}_{\bm n}, {\bm z}_{\bm x})$ in these models incorporates an image latent vector ${\bm z}_{\bm x}$ along with an noise latent vector ${\bm z}_{\bm n}$. The images were normalized within the range of $[-1, 1]$.

\smallskip\noindent\textbf{Training settings.}
As a GAN objective function, we used a non-saturating GAN loss~\cite{IGoodfellowNIPS2014} with $R_1$ regularization~\cite{LMeschederICML2018}. We set the trade-off parameter for $R_1$ regularization to $10$. In NR-GANs, we additionally used DS regularization~\cite{DYangICLR2019} with a trade-off parameter of 1.

During the experiments on \textsc{LSUN Bedroom},
we trained the networks for $500k$ iterations using the Adam optimizer~\cite{DPKingmaICLR2015} with $\alpha = 0.0001$, $\beta_1 = 0$, $\beta_2 = 0.99$, and a batch size of 64. We updated the generator and discriminator alternatively. Similar to prior studies~\cite{TKarrasICLR2018,LMeschederICML2018}, we used an exponential moving average with a decay of 0.999 over the weights to produce the final generator.

During the experiments on \textsc{FFHQ},
we trained the networks for $300k$ iterations, and the other settings were the same as those described in \textsc{LSUN Bedroom}.

\subsubsection{Evaluation details}
\label{subsubsec:detail_complex_evaluation}

For the same reason as that described in Appendix~\ref{subsubsec:detail_comprehensive_study_evaluation}, we used the FID as an evaluation metric. When calculating the FID, we used $10k$ real test samples and $10k$ generated samples, following the suggestion from previous large-scale studies on GANs~\cite{MLucicNeurIPS2018,KKurachICML2019}.

\subsection{Details on Section~\ref{subsec:application}}
\label{subsec:detail_application}

\subsubsection{Details of denoisers}
\label{subsubsec:detail_application_denoiser}

In this section, we describe the details of the denoisers (N2C, N2N, N2V, N2S, and GN2GC) described in Section~\ref{subsec:application}.

\smallskip\noindent\textbf{Network architecture.}
We used the same network architecture as that described in Appendix~\ref{subsubsec:detail_comprehensive_study_denoiser} (Table~\ref{tab:net_denoiser}). This is also the same as that used in the study on N2N~\cite{JLehtinenICML2018}. Note that this network is fully convolutional; therefore, it can take an arbitrary-size image as an input. The input images were normalized within the range of $[-0.5, 0.5]$.

\smallskip\noindent\textbf{Training settings.}
During the experiments on \textsc{LSUN Bedroom}, we trained the network for $500k$ iterations using the Adam optimizer~\cite{DPKingmaICLR2015} with $\alpha = 0.0003$, $\beta_1 = 0.9$, and $\beta_2 = 0.99$. The learning rate was kept constant during the training except for the last $30\%$ iterations, where the learning rate was smoothly ramped down to zero. Following the study on N2N~\cite{JLehtinenICML2018}, we used a batch size of 4 for N2N. For a fair comparison, we also used a batch size of 4 for N2C and GN2GC. This means that N2C, N2N, and GN2GC were trained under the same setting except for different input and output images. Similar to the observations discussed in Appendix~\ref{subsubsec:detail_comprehensive_study_denoiser}, we found that N2V and N2S operate better with a larger batch size. Hence, we used a batch size of 64 for N2V and N2S. Following the study on N2V~\cite{AKrullCVPR2019}, for N2V, we manipulated 64 pixels per input image and used a uniform pixel selection (UPS) with a kernel size of $5 \times 5$ as a masking method. With N2S, instead of UPS, we used random overwriting as a masking method~\cite{JBatsonICML2019}, i.e., the pixel is overwritten with a random color within the range of $[-0.5, 0.5]$.

During the experiments on \textsc{FFHQ}, we trained the network for $300k$ iterations, and the other settings were the same as those in \textsc{LSUN Bedroom}.

\subsubsection{Evaluation details}
\label{subsubsec:detail_application_evaluation}

As an evaluation metric, we used the PSNR, which is commonly used in image denoising. Herein, the score averaged over all test sets is provided.

\end{document}